\pdfoutput=1
\documentclass{article}

\usepackage{iclr2027_conference,times}

\usepackage{amsmath,amsfonts,bm}

\def\eqref#1{equation~\ref{#1}}

\def\1{\bm{1}}

\DeclareMathAlphabet{\mathsfit}{\encodingdefault}{\sfdefault}{m}{sl}
\SetMathAlphabet{\mathsfit}{bold}{\encodingdefault}{\sfdefault}{bx}{n}

\usepackage{amssymb}
\usepackage{pifont}
\usepackage{graphicx}
\usepackage{booktabs}
\usepackage{tabularx}
\usepackage{multirow}
\usepackage{array}
\usepackage{adjustbox}
\usepackage{float}
\usepackage{xcolor}
\usepackage{microtype}
\definecolor{iccvblue}{rgb}{0.21,0.49,0.74}
\usepackage[colorlinks,linkcolor=red,citecolor=iccvblue]{hyperref}
\usepackage{url}
\usepackage{wrapfig}
\newcolumntype{Y}{>{\centering\arraybackslash}X}
\newcolumntype{L}{>{\raggedright\arraybackslash}X}
\usepackage{colortbl}
\newcommand{\ablnum}[1]{{\color{black}\ding{\numexpr181+#1\relax}}}

\definecolor{VastMAT}{RGB}{224,244,247}

\newcommand{\figureplaceholder}[2]{%
  \IfFileExists{#1}{%
    \includegraphics[width=\linewidth]{#1}%
  }{%
    \fbox{\parbox[c][#2][c]{0.94\linewidth}{%
      \centering\ttfamily #1\\[0.5em]
      \normalfont\footnotesize Figure placeholder%
    }}%
  }%
}

\title{VastMAT: A Large-Scale Multi-Category Benchmark for Multi-Animal Tracking}

\author{%
\parbox[t]{\dimexpr\textwidth-2\tabcolsep\relax}{%
\centering
\bfseries
Zhizhen Li$^{1,2}$\thanks{Equal contribution. $\dagger$ Corresponding author.}
\quad
Zan Wang$^{3,*}$
\quad
Huidong Peng$^{2,*}$
\quad
Bohan Tan$^{4,*}$
\quad
Shimin Shan$^{1}$
\quad \\[2pt]
Yu Liu$^{1}$
\quad
Liang Peng$^{2,\dagger}$
\\[3pt]
\normalfont
$^{1}$Dalian University of Technology
\qquad
$^{2}$Wuhan University
\qquad
$^{3}$University of North Texas
\qquad
$^{4}$The Hong Kong University of Science and Technology
\\[3pt]
\texttt{lizhizhen@mail.dlut.edu.cn}, \quad \texttt{pengliang@whu.edu.cn}
}}

\iclrfinalcopy 

\begin{document}

\maketitle
\fancyhead{}
\renewcommand{\headrulewidth}{0pt}

\begin{abstract}
Multi-animal tracking (MAT) supports the study of animal movement, behavior, and group interactions.
However, general multi-object tracking (MOT) benchmarks primarily focus on pedestrians and vehicles, whereas dedicated MAT benchmarks remain limited in jointly supporting broad animal coverage, large-scale video data, and extensive within-video multi-instance association.
To address this gap, we introduce \textbf{VastMAT}, which has four key characteristics:
\textbf{(1) Large scale.} It comprises 2,947 videos with 1,002,562 annotated frames, totaling 27.85 hours.
\textbf{(2) Broad category coverage.} These videos cover 337 animal categories with diverse morphologies and motion patterns.
\textbf{(3) Extensive instance annotations.} It provides 3,663,248 bounding boxes and 22,883 identity trajectories---to our knowledge, the largest numbers of both among dedicated MAT benchmarks.
\textbf{(4) High-quality annotations.} To ensure reliability, annotations undergo iterative expert review and correction, and quality is assessed through an independent reannotation audit.
To systematically assess tracking performance and cross-category generalization, we establish Seen-category and category-disjoint Unseen-category protocols, and evaluate eight representative MOT methods under both protocols.
Under these protocols, the highest baseline HOTA scores are 66.37\% and 52.90\%, respectively, highlighting the challenge of tracking unseen animals.
To address the low-overlap association challenge revealed by our analysis, we propose Center-Distance-Augmented Association (CDA), a lightweight module that adaptively combines IoU with center similarity normalized by the boxes' own scales.
Without additional training, CDA improves TrackTrack's HOTA by 1.58 and 1.31 percentage points under the two protocols, respectively. To facilitate further MAT research, we will publicly release our benchmark and code.
\end{abstract}

\section{Introduction}
\label{sec:introduction}

\begin{wrapfigure}{r}{0.5\textwidth}
  \centering
  \vspace{-0.2cm}
  \figureplaceholder{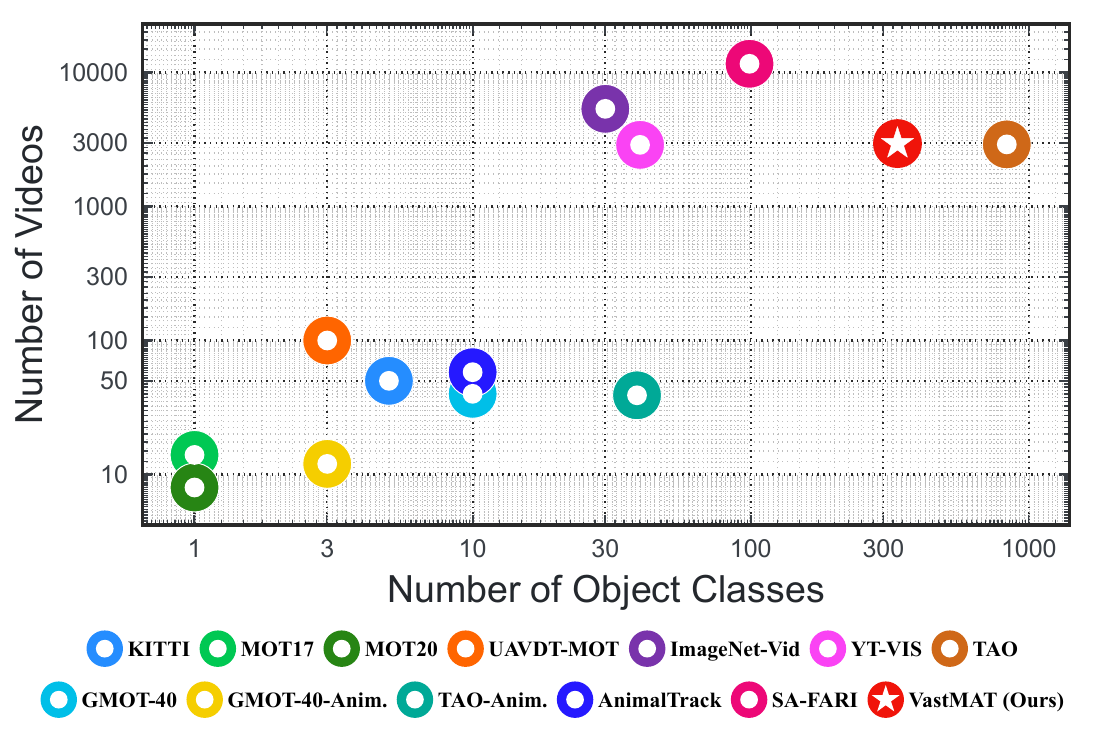}{\linewidth}
  \caption{Category and video counts of VastMAT and representative tracking datasets.}
  \vspace{-0.2cm}
  \label{Figdataset_bijiao}
\end{wrapfigure}

Multi-animal tracking (MAT) is a branch of multi-object tracking (MOT) that localizes all visible animals in a video and maintains their identities across frames. The resulting identity-consistent trajectories support studies of animal movement, individual behavior, and group interactions, making reliable MAT valuable to biology, ecology, conservation, and animal husbandry. Yet research on tracking across diverse animal categories remains constrained by the scale, species diversity, and association density of existing benchmarks.

Evaluating tracking across diverse animals requires attention to both category variation and within-video identity association. Differences in morphology, motion, and imaging conditions demand benchmarks that test applicability across animal categories, including Unseen animals. Animal videos also range from a few independently moving individuals to dense interactions among similar-looking animals, where rapid displacement, non-rigid deformation, and occlusion complicate identity preservation. MAT benchmarks therefore need both continuous multi-instance identity annotations and complementary evaluations that distinguish these capabilities.

Existing MOT benchmarks primarily target pedestrians or vehicles, including MOT17, MOT20, KITTI, and UAVDT \citep{dendorfer2021motchallenge,dendorfer2020mot20,geiger2012kitti,du2018uav}. DanceTrack and SportsMOT introduce similar appearances and complex motion but remain focused on humans \citep{sun2022dancetrack,cui2023sportsmot}. TAO, BURST, ImageNet-Vid, and YouTube-VIS broaden category and task coverage, yet animals constitute only part of these datasets, whose annotation frequencies, output formats, and evaluation protocols differ\citep{dave2020tao,athar2023burst,russakovsky2015imagenet,yang2019ytvis}. A large general-purpose vocabulary thus does not necessarily provide broad coverage of animals, group interactions, or multi-instance association across animal categories.

Dedicated animal tracking benchmarks have advanced MAT but still struggle to combine video scale, category diversity, and multi-instance identity association. AnimalTrack provides 58 videos from 10 animal categories for association in dense animal groups, but its category and video coverage remain limited\citep{zhang2023animaltrack}. SA-FARI expands coverage to 11,609 camera-trap videos and 99 animal categories with pixel-level annotations, yet its average of approximately 1.4 identity trajectories per video differs from the association demands of dense group tracking\citep{wasmuht2026safari}. MAT therefore needs a unified benchmark combining diverse animal categories, large-scale video data, and continuous multi-instance identity annotations.

\begin{figure}[t]
    \centering
    \figureplaceholder{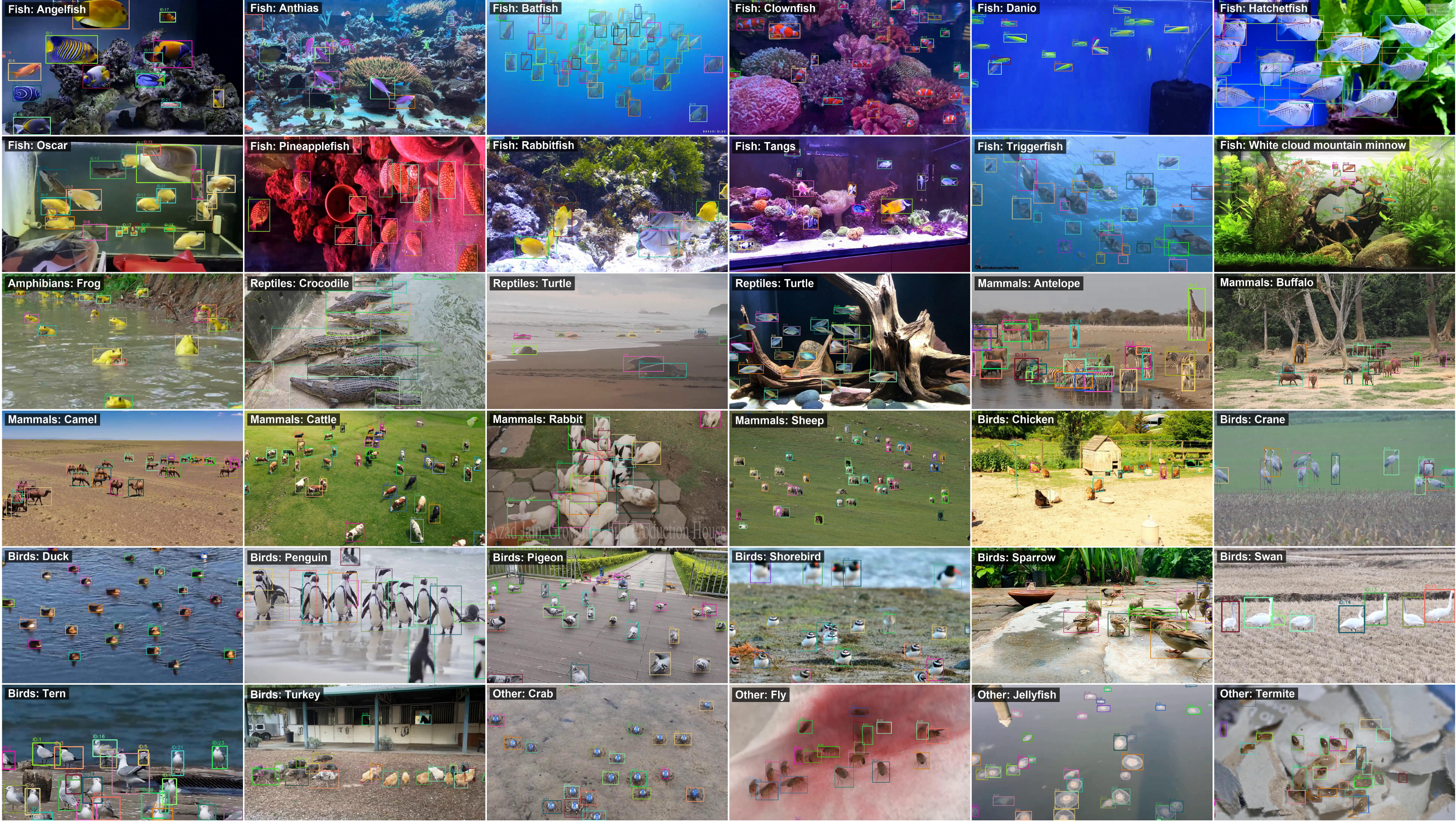}{0.31\textheight}
    \caption{Annotated examples across animal groups and scenes in VastMAT.}
    \label{fig:overview}
\end{figure}

To this end, we introduce VastMAT, comprising 337 animal categories, 2,947 videos, 1,002,562 annotated frames, 3,663,248 bounding boxes, and 22,883 identity trajectories over 27.85 hours. As Figure~\ref{fig:overview} illustrates, the dataset covers fish, birds, mammals, reptiles, amphibians, and other animals, with individual motion, category co-occurrence, and group interactions in underwater, terrestrial, aerial, and artificial environments. Videos vary in object scale, viewpoint, lighting, background, and camera motion, and present association challenges involving deformation, occlusion, and similar appearances. Figure~\ref{Figdataset_bijiao} and Table~\ref{tab:dataset_comparison} compare category coverage and dataset scale. With 7.76 trajectories per video on average, VastMAT combines broad category coverage with continuous multi-instance association to support tracking and Cross-category generalization evaluation.

\begin{table}
    \vspace*{-0.6em}
    \caption{Detailed comparison of VastMAT and representative tracking datasets. n/a indicates that the source does not provide a directly comparable value. GMOT-40-Anim. and TAO-Anim. are animal subsets recomputed from their parent datasets.}
    \label{tab:dataset_comparison}
    \centering
    \footnotesize
    \setlength{\tabcolsep}{2.5pt}
    \setlength{\aboverulesep}{0pt}
    \setlength{\belowrulesep}{0pt}
    \renewcommand{\arraystretch}{1.1}
    \begin{adjustbox}{width=\linewidth}
    \begin{tabular}{r *{13}{c}}
    \toprule
    \rowcolor{gray!12}
        \raisebox{1.5ex}{Benchmark}
        & \raisebox{1.5ex}{Videos}
        & \raisebox{1.5ex}{Categories}
        & \shortstack[c]{Min.\\len.(s)} & \shortstack[c]{Avg.\\len.(s)} & \shortstack[c]{Max.\\len.(s)} & \shortstack[c]{Total\\len.(s)} & \shortstack[c]{Avg.\\tracks} & \shortstack[c]{Max.\\tracks} & \shortstack[c]{Total\\tracks} & \shortstack[c]{Frame\\rate} & \shortstack[c]{Ann.\\FPS} & \shortstack[c]{Total\\boxes} & \shortstack[c]{Total\\frames} \\
        \midrule
        KITTI~\citep{geiger2012kitti} & 50 & 5 & n/a & 10 & n/a & 498 & 52 & n/a & 2,600 & 30 & 10 & 80K & 15K \\
        MOT17~\citep{dendorfer2021motchallenge} & 14 & 1 & 17 & 33 & 85 & 463 & 95 & 222 & 1,331 & 25 & 30 & 300K & 11K \\
        MOT20~\citep{dendorfer2020mot20} & 8 & 1 & 17 & 66.8 & 133 & 535 & 479 & 1,211 & 3,833 & 25 & 30 & 2,102K & 13K \\
        UAVDT-MOT~\citep{du2018uav} & 100 & 3 & 2.8 & 26.67 & 99 & 2,666.7 & 27.0 & n/a & 2,700 & 30 & 6 & 840K & 40K \\
        ImageNet-Vid~\citep{russakovsky2015imagenet} & 5,354 & 30 & 0.2 & 12.1 & 219.7 & 64,547.9 & n/a & n/a & n/a & 25 & 25 & n/a & 1,614K \\
        YT-VIS~\citep{yang2019ytvis} & 2,883 & 40 & 1 & 5.6 & 7.2 & 16,015.2 & n/a & n/a & n/a & 30 & 5 & 131K & 480.5K \\
        TAO~\citep{dave2020tao} & 2,907 & 833 & n/a & 36.8 & n/a & 106,978 & 6 & 10 & 17,287 & 30 & 1 & 333K & 2,674K \\
        GMOT-40~\citep{bai2021gmot40} & 40 & 10 & 3 & 8.9 & 24.2 & 356.0 & 51 & 128 & 2,026 & 30 & 30 & 256K & 9K \\
        GMOT-40-Anim.~\citep{bai2021gmot40} & 12 & 3 & 3 & 7.1 & 24.2 & 85.5 & 70 & 128 & 837 & 30 & 30 & 63K & 2.6K \\
        TAO-Anim.~\citep{dave2020tao} & 39 & 39 & 1 & 22 & 93 & 859.0 & 4 & 10 & 250 & 30 & 1 & 3.4K & 2.5K \\
        AnimalTrack~\citep{zhang2023animaltrack} & 58 & 10 & 6.5 & 14.2 & 75.6 & 823.7 & 33 & 133 & 1,927 & 30 & 30 & 429K & 24.7K \\
        SA-FARI~\citep{wasmuht2026safari} & 11,609 & 99 & 0.5 & 14.19 & 90 & 164,808 & 1.4 & 14 & 16,224 & 10--60 & 6 & 942K & 988K \\
        \midrule
        \rowcolor{VastMAT}
        \textbf{VastMAT}(Ours) & 2,947 & 337 & 4 & 34.02 & 100.5 & 100,256.2 & 7.76 & 152 & 22,883 & 10 & 10 & 3,663K & 1,003K \\
        \bottomrule
    \end{tabular}
    \end{adjustbox}
\end{table}

We establish two protocols to evaluate tracking on new videos of Seen categories and generalization to Unseen animal categories. Each contains 2,632 training videos and 315 test videos. The Unseen-category protocol strictly separates 283 training categories from 54 test categories, assigning categories that co-occur in a video to the same split. We systematically evaluate eight representative MOT methods. DiffMOT achieves the highest baseline HOTA under both protocols, scoring 66.37 and 52.90, while the corresponding YOLOX-X detectors achieve animal detection AP scores of 72.34 and 54.07. Together with animal-group and video-level analyses, these results show that evaluating tracking across diverse animals requires examining object detection, identity association, and performance variation across scenes.


We further introduce Center-Distance-Augmented Association (CDA), a lightweight, training-free module that replaces the geometric matching score in existing trackers at inference time. CDA combines IoU with center similarity normalized by the mean box diagonal and adapts their weights to the number of high-confidence detections. It improves TrackTrack's HOTA by 1.58 and 1.31 percentage points under the Seen- and Unseen-category protocols, respectively.



In summary, Our main contributions are as follows: \ding{171} We establish VastMAT, a large-scale MAT benchmark with 2,947 videos and 337 animal categories, and validate box-level and within-clip identity consistency through an independent reannotation audit. \ding{170} We define Seen-category and category-disjoint Unseen-category protocols, systematically evaluate eight MOT methods, and analyze tracking through detection, animal-group, and video-level results. \ding{168} We propose lightweight CDA to improve identity association through normalization by the boxes' own scales and adaptive similarity fusion, and validate its effectiveness through geometric comparisons, fusion-weight analyses, and cross-tracker experiments.

\section{Related Work}
\label{sec:related_work}
\vspace{-5pt}

\textbf{Multi-object tracking benchmarks.}
MOT17, MOT20, KITTI, and UAVDT-MOT provide pedestrian and vehicle tracking evaluations~\citep{dendorfer2021motchallenge,dendorfer2020mot20,geiger2012kitti,du2018uav}. DanceTrack and SportsMOT introduce complex motion and similar appearances but remain human-focused~\citep{sun2022dancetrack,cui2023sportsmot}. ImageNet-Vid, YouTube-VIS, TAO, and BURST broaden category coverage across detection, segmentation, and tracking~\citep{russakovsky2015imagenet,yang2019ytvis,dave2020tao,athar2023burst}, while GMOT-40 evaluates exemplar-guided generic tracking~\citep{bai2021gmot40}. Their task definitions, annotation frequencies, and animal coverage differ, leaving limited support for continuous identity association across diverse animal groups.

\textbf{Animal vision and tracking benchmarks.}
Animal Kingdom, MammalNet, and ChimpACT support animal behavior analysis~\citep{ng2022animalkingdom,chen2023mammalnet,ma2023chimpact}; AP-10K and APT-36K provide cross-species pose annotations~\citep{yu2021ap10k,yang2022apt36k}; FishNet supports both fish recognition and detection, while Wildlife-71 and WildlifeReID-10k address individual re-identification~\citep{khan2023fishnet,jiao2023animalreid,adam2025wildlifereid}. Together, These resources primarily target behavior, pose, recognition, or image-level identity. Among MAT benchmarks, AnimalTrack emphasizes dense groups but has limited category and video coverage~\citep{zhang2023animaltrack}. SA-FARI broadens coverage with pixel-level annotations but provides relatively sparse within-video identities~\citep{wasmuht2026safari}. VastMAT combines 337 animal categories and 2,947 videos with dense frame-by-frame boxes and persistent identities for multi-instance association and cross-category evaluation.

\textbf{Multi-object tracking and geometric association.}
Tracking-by-detection typically combines motion and appearance cues to establish identity correspondences. ByteTrack associates high- and low-confidence detections to recover overlooked objects\citep{zhang2022bytetrack}, DiffMOT predicts nonlinear motion using diffusion models\citep{lv2024diffmot}, and TrackTrack improves candidate matching and track initialization from a track-centric perspective\citep{shim2025TrackTrack}.
Joint approaches, such as TransTrack and MOTIP, maintain identities through track queries and temporal identity prediction\citep{sun2020transtrack,gao2025motip}. In explicit candidate matching, geometric scores measure spatial compatibility between predicted and detected boxes. GIoU, DIoU, CIoU, and EIoU extend IoU using enclosing regions, center distance, or shape information\citep{rezatofighi2019generalized,zheng2020diou,zhang2022focal}. To address low-overlap association caused by animal motion and deformation, CDA combines IoU with linear center similarity normalized by the boxes' own scales and adjusts their weights to the detection count, requiring no additional training.

\section{The VastMAT Dataset}
\label{sec:dataset}

\subsection{Design Principles}
\label{sec:design_principles}

VastMAT provides a unified, large-scale resource for multi-category MAT, animal detection, and Cross-category generalization. Its construction follows four principles. \textbf{(i) Broad animal coverage.}
    We aim to cover at least 300 animal categories with diverse morphologies and motion patterns across underwater, terrestrial, and aerial environments.
    \textbf{(ii) Large scale.}
    We expand video, annotated-frame, bounding-box, and trajectory counts to capture diverse imaging conditions and support training and evaluation.
    \textbf{(iii) Extensive identity association.}
    We retain videos with multiple animals, co-occurring categories, and dense interactions to test identity preservation under occlusion, entry and exit, and similar appearances.
    \textbf{(iv) Reliable annotations.}
    We standardize category labels, bounding boxes, and identity assignment, combining model-assisted annotation with repeated manual review and independent reannotation to validate consistency.

\subsection{Data Collection}
\label{sec:data_collection}

We curate animal names, synonyms, and potentially confusable categories, selecting 337 animal categories from more than 400 candidates (Appendix~\ref{app:animal_grouping}). Experts, including doctoral and master's students working on related topics, verify each category's suitability for tracking. We then search YouTube for Creative Commons-licensed videos of each category, collecting more than 5,000 candidate clips. After reviewing their suitability for visual tracking, we retain 2947 sequences and sample them uniformly at 10 FPS. Figure~\ref{Fig_datasetchangwei} in the appendix shows the long-tailed distributions of videos, trajectories, and boxes across categories, reflecting their differing frequencies in public video sources. Appendix~\ref{app:maintenance_ethics} details provenance, maintenance, and responsible use.

\subsection{Annotation and Quality Validation}
\label{sec:annotation_quality_format}

We annotate animal boxes, categories, and within-video identities using X-AnyLabeling 3.3.7 and SAM3~\citep{wang2023xanylabeling,carion2025sam3}. Each video is assigned to one annotator, who initializes animal regions, propagates them with SAM3, and corrects missed objects, drift, and identity errors. Boxes cover visible animal regions; fully occluded animals are omitted. Reappearing animals retain their IDs when identity is verifiable and receive new IDs otherwise. Second, experts review object coverage, localization, categories, and identity continuity. Third, annotations without unanimous approval from the two- to three-expert review team are returned to the original annotator for correction. Review and correction are repeated over multiple rounds. Finally, masks are converted to AnimalTrack-compatible bounding boxes. Table~\ref{tab:gt_format} in the appendix defines the format, and Figure~\ref{fig:overview} shows examples.

We audit annotation quality by randomly sampling 48 of the 2947 videos across all six animal groups and selecting 2 non-overlapping, consecutive 20-frame clips per video, totaling 1,920 frames. Annotators trained under the same guidelines, but uninvolved in the original annotation and without access to it, independently reannotate these frames, followed by the same repeated review and correction stages. We then match the 1920 reannotated frames one-to-one with the originals. At $\mathrm{IoU}\geq0.5$, box-matching F1 is $96.03\%$, and mean IoU among matched boxes is $0.9555$. All 7,632 matched box pairs agree in category. To assess temporal identity consistency, we check whether each original identity matched in at least two frames of a 20-frame clip always maps to the same reannotated identity. Of 427 eligible clip--identity pairs, 424 are consistent, yielding $99.30\%$ within-clip identity consistency and a mean longest consecutive consistent span of $18.49$ frames. These results support the reliability of object coverage, localization, and within-clip identities. Appendix~\ref{app:annotation_quality} provides full metrics and group-wise results.

\subsection{Dataset Statistics and Characteristics}

VastMAT provides 27.85 hours of annotations across 337 animal categories, averaging 7.76 trajectories per video and 3.65 objects per frame. Average trajectory and box counts in Table~\ref{tab:split_statistics} are both computed per video. The 337 categories comprise 82 mammals, 85 birds, 117 fish, 4 amphibians, 7 reptiles, and 42 other animals; Appendix~\ref{app:animal_grouping} lists all categories and their protocol coverage.

VastMAT includes 607 videos with more than 10 identity trajectories and 827 multi-category videos, covering sparse individual motion, multi-instance interactions, and category co-occurrence. Appendix~\ref{app:dataset_statistics} provides further distribution and co-occurrence analyses, while Figure~\ref{Figxilidu} illustrates identity discrimination challenges involving visually similar categories.

\begin{figure*}
    \centering
    \figureplaceholder{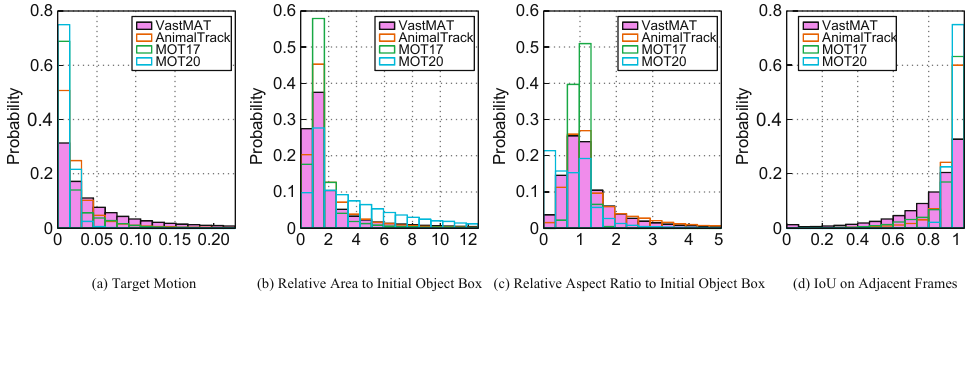}{\textwidth}
    \vspace{-1.5cm}
    \caption{Motion and box-geometry distributions across tracking datasets.}
    \label{fig:temporal_geometric_variation}
\end{figure*}

\begin{wraptable}{r}{0.5\textwidth}
    \vspace*{-0.3em}
    \caption{IoU statistics by normalized displacement (Protocol~2). Event percentages may not sum to 100\% due to rounding.}
    \label{tab:speed_bucket_motivation}
    \centering
    \footnotesize
    \setlength{\tabcolsep}{2.5pt}
    \setlength{\aboverulesep}{0pt}
    \setlength{\belowrulesep}{0pt}
    \renewcommand{\arraystretch}{1.1}
    \begin{adjustbox}{width=\linewidth}
    \begin{tabular}{lccccc}
    \toprule
    \rowcolor{gray!12}
    Displ. & Events & IoU=0 & $<0.1$ & $<0.3$ & Median \\
    \midrule
    \midrule
    $\leq 0.25$        & 95.47\% & 0.004\% & 0.020\% & 0.304\% & 0.894 \\
    $0.25$--$0.5$      & 3.25\%  & 1.49\%  & 5.58\%  & 51.19\% & 0.296 \\
    $0.5$--$1.0$       & 0.94\%  & 29.49\% & 71.62\% & 100.0\% & 0.050 \\
    $>1.0$             & 0.35\%  & 100.0\% & 100.0\% & 100.0\% & 0.000 \\
    \bottomrule
    \end{tabular}
    \end{adjustbox}
\end{wraptable}

Figure~\ref{fig:temporal_geometric_variation} compares motion, relative area, relative aspect ratio, and consecutive-frame IoU for the same identity. Motion and IoU are computed between consecutive annotated frames, while area and aspect ratio are normalized to the first-frame box. VastMAT has a median Target Motion of 0.032, compared with 0.015, 0.008, and 0.010 for AnimalTrack, MOT17, and MOT20. Its 95th percentile is 0.284, approximately 3.0, 3.7, and 10.6 times the respective values. The fraction with IoU$<0.8$ is $33.4\%$, compared with $8.6\%$ for AnimalTrack. Together with broader area and aspect-ratio distributions, these results indicate greater temporal geometric variation in VastMAT.

Protocol~2 contains 278,390 valid GT box pairs with the same identity and a frame-index gap of at most 3. We define normalized displacement as center distance divided by the mean box diagonal. As Table~\ref{tab:speed_bucket_motivation} shows, median IoU is 0.894 for displacements up to 0.25, falling to 0.296 and 0.050 for 0.25--0.5 and 0.5--1.0, respectively. Zero overlap accounts for only $0.68\%$ of all events; low overlap is concentrated in the large-displacement ranges.

\subsection{Data Splits and Evaluation Protocols}
\label{sec:protocols}

\textbf{Data splits.}
We refer to the data partitioning principle in \citep{ying2025move}. Both protocols contain 2,632 training videos and 315 test videos but use different splits (See Table~\ref{tab:split_statistics}). All Protocol~1 test categories appear in the training set, whereas Protocol~2 strictly separates 283 training categories from 54 test categories. To keep co-occurring categories on the same side of the split, we assign entire connected components of the category co-occurrence graph, placing the largest component of 266 categories in training.

\begin{table}[t]
    \vspace*{-0.4em}
    \caption{Split statistics for the two VastMAT evaluation protocols.}
    \label{tab:split_statistics}
    \centering
    \footnotesize
    \setlength{\tabcolsep}{2.5pt}
    \setlength{\aboverulesep}{0pt}
    \setlength{\belowrulesep}{0pt}
    \renewcommand{\arraystretch}{1.1}
    \begin{adjustbox}{width=\linewidth}
    \begin{tabular}{l*{10}{c}}
        \toprule
        \rowcolor{gray!12}
        Split & Videos & Classes & Min. (s) & Avg. (s) & Max. (s) &
        Frames & Tracks & Boxes & Avg. tracks & Avg. boxes \\
        \midrule
        \midrule
        Exp1 Train &
        2,632 & 337 & 4.0 & 34.556 & 100.5 &
        909,526 & 20,782 & 3,389,474 & 7.90 & 1,287.79 \\

        Exp1 Test &
        315 & 316 & 5.1 & 29.535 & 100.0 &
        93,036 & 2,101 & 273,774 & 6.67 & 869.12 \\

        Exp2 Train/Seen &
        2,632 & 283 & 4.0 & 34.554 & 100.5 &
        909,451 & 20,734 & 3,351,099 & 7.88 & 1,273.21 \\

        Exp2 Test/Unseen &
        315 & 54 & 7.4 & 29.559 & 100.0 &
        93,111 & 2,149 & 312,149 & 6.82 & 990.95 \\
        \bottomrule
    \end{tabular}
    \end{adjustbox}
\end{table}

\textbf{Evaluation protocols.}
For both detection and tracking, all valid animal objects are mapped to a single animal foreground class. Table~\ref{tab:split_statistics} summarizes the splits. Protocol~2 test categories cover mammals, birds, fish, and other animals; all amphibians and reptiles remain in training and are evaluated only under Protocol~1. Table~\ref{tab:animal_group_coverage} in the appendix provides the full composition. Because training-category coverage and test-video composition both differ between protocols, cross-protocol gaps reflect their combined effects.

\begin{wrapfigure}{r}{0.5\textwidth}
    \centering
    \vspace{-0.3cm}
    \includegraphics[width=0.49\textwidth]{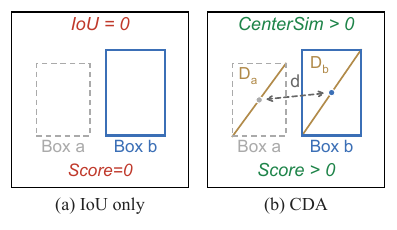}
    \vspace{-0.1cm}
    \caption{Scoring non-overlapping candidates with (a) IoU and (b) CDA.}
    \label{fig:iou_failure}
    \vspace{-0.4cm}
\end{wrapfigure}

\section{CDA: A Lightweight Association Module}
\label{sec:method}

\subsection{Motivation and Overview}
\label{sec:method_motivation}

CDA is a lightweight, training-free module for geometric association at inference time. Rapid motion and deformation can produce low-overlap candidates(See Table~\ref{tab:speed_bucket_motivation}). Figure~\ref{fig:iou_failure} illustrates the zero-overlap case: nearby boxes retain positive CenterSim when their center distance is smaller than the mean box diagonal, yielding a positive fused score for $\alpha<1$. CDA combines this cue with IoU, normalizes center distance by object scale, and adapts fusion weights to detection count.

\subsection{CDA Similarity Score}
\label{sec:method_centersim}

Given a track's predicted box $a$ and a candidate detection box $b$ in the current frame, CDA defines the fused similarity as
\begin{equation}
S(a,b)=
\alpha\,\mathrm{IoU}(a,b)
+(1-\alpha)\,\mathrm{CenterSim}(a,b),
\label{eq:similarity}
\end{equation}
where $\alpha$ controls the relative contributions of region overlap and center proximity. Center similarity is defined as
\begin{equation}
\mathrm{CenterSim}(a,b)=
\mathrm{clip}\left(
1-\frac{d_{\mathrm{center}}}
{(\mathrm{diag}_a+\mathrm{diag}_b)/2},
\,0,\,1
\right).
\label{eq:centersim}
\end{equation}
Here, $d_{\mathrm{center}}$ is the Euclidean distance between box centers, and $\mathrm{diag}_a$ and $\mathrm{diag}_b$ are their diagonal lengths. CenterSim equals 1 when the centers coincide and falls to 0 when their distance reaches the mean diagonal length. Thus, some spatially close but non-overlapping candidates retain nonzero similarity, supplying positional information when overlap is insufficient. When pose changes shift box boundaries while the centers remain close, CenterSim complements overlap; IoU retains region-level spatial constraints that help distinguish competing candidates with similar center distances.

\subsection{Box-Scale Normalization}
\label{sec:method_diou}

The same pixel displacement has different implications for animals of different sizes. Equation~(\ref{eq:centersim}) normalizes center distance by the mean box diagonal, making CenterSim invariant to uniform scaling of boxes and their positions. Averaging incorporates predicted and detected scales, so the distance reference is not determined solely by the smaller box. CDA draws on DIoU's use of center distance to complement overlap\citep{zheng2020diou}, but uses the boxes' scales and linear decay. Unlike the diagonal of the smallest enclosing rectangle, this denominator does not grow as the enclosing region expands with center separation, preserving a distance scale determined by object size. Section~\ref{sec:exp_cda} compares geometric designs; Appendix~\ref{app:cda_gate} defines normalized DIoU and candidate gating.

\subsection{Detection-Count-Adaptive Fusion}
\label{sec:method_density}

With few animals, center proximity helps compensate for low overlap; dense scenes require stronger overlap constraints to distinguish competing objects. We approximate group size using the number of high-confidence detections $n$ in the current frame and define
\begin{equation}
\alpha(n)=
\mathrm{clip}\left(
0.5+k(n-N_{\mathrm{ref}}),\,0.1,\,0.9
\right).
\label{eq:alpha}
\end{equation}
The defaults are $k=0.05$ and $N_{\mathrm{ref}}=10$.
The IoU weight is 0.1, 0.5, and 0.9 for $n\leq2$, $n=10$, and $n\geq18$, respectively. Restricting it to $[0.1,0.9]$
keeps both geometric cues active in the adaptive configuration. The rule increases the IoU contribution as detection count grows and favors center similarity when fewer objects are present.

CDA replaces the base tracker's geometric association term: it computes $S$ from predicted and current-frame detection boxes and uses $1-S$ as the geometric cost. In TrackTrack, this cost is combined with the existing appearance, confidence, and direction terms; the remaining matching and track-update procedures are unchanged. CDA and the DIoU control use identical normalized-DIoU gating to compare scores for admitted candidates; Section~\ref{sec:cross_tracker} discusses changes in candidate acceptance in ByteTrack. CDA requires no additional training. Default parameters are selected on a development set drawn from the training split and fixed during testing. Appendix~\ref{app:cda_details} provides the full association cost and implementation details.

\section{Experiments}
\label{sec:experiments}

\noindent\textbf{Evaluation metrics.}
We use TrackEval to compute HOTA, CLEAR, and Identity metrics
\citep{luiten2021hota,bernardin2008clear,ristani2016id}.
HOTA jointly evaluates detection and association through DetA and AssA, which we also report in diagnostic experiments. MOTA aggregates false negatives, false positives, and identity switches, while IDF1 measures identity matching consistency. We additionally report IDP/IDR, MT/PT/ML, FP/FN, identity switches (IDs/IDSW), and track fragmentations (FM). Detection is evaluated using COCO-style AP, AP$_{50}$, and AP$_{75}$, with AP averaged over IoU thresholds from 0.50--0.95 in steps of 0.05.

\subsection{Evaluated Methods}
\label{sec:evaluated_trackers}

We evaluate the eight trackers in Table~\ref{tab:Protocol_1+2_results} under both protocols. They span motion matching, appearance representation, track queries, and temporal identity modeling, enabling comparison of these mechanisms in animal scenes. ByteTrack, DiffMOT, and TrackTrack use YOLOX-X trained on the corresponding protocol's training split, followed by their respective association pipelines. End-to-end methods train their joint detection and association frameworks on the corresponding training split.

\subsection{Evaluation Results}
\label{sec:evaluation_results}

\begin{wraptable}{r}{0.31\textwidth}
    \centering
    \vspace{-0.1cm}
    \caption{Overall detection metrics and group-wise AP (\%) for YOLOX-X.}
    \label{tab:detection_overall}
    \footnotesize
    \setlength{\tabcolsep}{4pt} 
    \setlength{\aboverulesep}{0pt}
    \setlength{\belowrulesep}{0pt}
    \renewcommand{\arraystretch}{1.1}
    \begin{adjustbox}{width=\linewidth}
    \begin{tabular}{lcc} 
        \toprule
        \rowcolor{gray!12}
        Metric / Group & P1 & P2 \\
        \midrule
        \midrule
        Overall AP       & 72.34 & 54.07 \\
        Overall AP\(_{50}\) & 84.22 & 67.47 \\
        Overall AP\(_{75}\) & 75.94 & 56.63 \\
        \midrule
        Mammals AP    & 74.38 & 54.62 \\
        Fish AP       & 72.68 & 68.36 \\
        Birds AP      & 75.77 & 74.67 \\
        Amphibians AP & 78.33 & --    \\
        Reptiles AP   & 25.24 & --    \\
        Others AP     & 45.98 & 23.08 \\
        \bottomrule
    \end{tabular}
    \end{adjustbox}
    \vspace{-0.4cm}
\end{wraptable}

\paragraph{Overall performance.}
Table~\ref{tab:Protocol_1+2_results} shows that DiffMOT achieves the highest HOTA under both protocols, at 66.37 and 52.90, followed by TrackTrack at 65.10 and 52.61. TrackTrack also produces the fewest track fragmentations in both. All eight methods score lower on Protocol~2, with HOTA gaps of approximately 5.3--13.5 percentage points, indicating a substantial performance gap under the Unseen-category evaluation setting. MOTIP and TCEI produce more mostly tracked trajectories in Protocol~1 but also more false positives. In Protocol~2, TCEI achieves the highest IDF1 of 56.27,
but its HOTA remains below DiffMOT and TrackTrack, underscoring the need to consider detection, identity association, and error statistics together.

\begin{table}[t]
    \vspace*{-0.4em}
    \caption{Overall comparison of tracking algorithms on VastMAT under Protocols 1 and 2. Within each protocol, the best two results for each metric are highlighted in \textcolor{red}{red} and \textcolor{blue}{blue}, respectively.}
    \label{tab:Protocol_1+2_results}
    \centering
    \footnotesize
    \setlength{\tabcolsep}{2pt}    
    \setlength{\aboverulesep}{0pt}
    \setlength{\belowrulesep}{0pt}
    \renewcommand{\arraystretch}{1.1}
    \begin{adjustbox}{width=\linewidth}
    \begin{tabular}{rcccccccccccc}
        \toprule
        \rowcolor{gray!12}
        Tracker & HOTA & MOTA & IDF1 & IDP & IDR & MT & PT & ML$\downarrow$ & FP$\downarrow$ & FN$\downarrow$ & IDs$\downarrow$ & FM$\downarrow$ \\
        \midrule
        \midrule
        \multicolumn{13}{c}{\textit{Protocol 1}} \\
        \midrule
        TransTrack \citep{sun2020transtrack} [arXiv 2020] & 41.33 & 39.05 & 44.33 & 54.84 & 37.19 & 519 & \textcolor{blue}{1029} & 553 & 37200 & 125298 & 4363 & 11599 \\
        QDTrack \citep{pang2021qdtrack} [CVPR 2021]       & 53.80 & 52.79 & 60.01 & 74.92 & 50.05 & 623 & 886  & 592 & 18419 & 109312 & \textcolor{red}{1523} & 8051 \\
        FairMOT \citep{zhang2021fairmot} [IJCV 2021]      & 41.31 & 47.23 & 43.93 & 53.43 & 37.30 & 513 & \textcolor{red}{1038} & 550 & 24279 & 106904 & 13300 & 10728 \\
        ByteTrack \citep{zhang2022bytetrack} [ECCV 2022]  & 56.59 & 62.91 & 63.37 & 69.93 & 57.93 & 846 & 905  & 350 & 25824 & 72789  & 2934 & 5903 \\
        DiffMOT \citep{lv2024diffmot} [CVPR 2024]         & \textcolor{red}{66.37} & \textcolor{red}{68.38} & \textcolor{red}{69.21} & \textcolor{red}{77.44} & 62.56 & 1043 & 710 & 348 & \textcolor{red}{15915} & 68519 & 2142 & \textcolor{blue}{5054} \\
        MOTIP \citep{gao2025motip} [CVPR 2025]            & 61.71 & 58.55 & 65.96 & 67.63 & \textcolor{blue}{64.36} & \textcolor{red}{1179} & 700 & \textcolor{blue}{222} & 48268 & \textcolor{blue}{61505} & 3712 & 7586 \\
        TrackTrack \citep{shim2025TrackTrack} [CVPR 2025] & \textcolor{blue}{65.10} & \textcolor{blue}{67.45} & \textcolor{blue}{67.87} & \textcolor{blue}{75.88} & 61.39 & 934 & 806  & 361 & \textcolor{blue}{17428} & 69720 & \textcolor{blue}{1962} & \textcolor{red}{4200} \\
        TCEI \citep{guo2026tcei} [CVPR 2026]              & 61.89 & 58.95 & 66.28 & 67.72 & \textcolor{red}{64.91} & \textcolor{blue}{1176} & 706 & \textcolor{red}{219} & 48668 & \textcolor{red}{60034} & 3672 & 7629 \\
        \midrule
        \multicolumn{13}{c}{\textit{Protocol 2}} \\
        \midrule
        TransTrack \citep{sun2020transtrack} [arXiv 2020] & 36.00 & 28.25 & 38.49 & 53.33 & 30.11 & 357 & \textcolor{red}{1116} & 676 & 41788 & 177662 & 4504 & 14274 \\
        QDTrack \citep{pang2021qdtrack} [CVPR 2021]       & 41.64 & 35.23 & 45.56 & 69.83 & 33.80 & 371 & 942  & 836 & 19569 & 180682 & \textcolor{red}{1928} & 9606 \\
        FairMOT \citep{zhang2021fairmot} [IJCV 2021]      & 32.85 & 34.33 & 34.82 & 49.60 & 26.83 & 335 & \textcolor{blue}{1104} & 710 & 24095 & 167431 & 13474 & 12348 \\
        ByteTrack \citep{zhang2022bytetrack} [ECCV 2022]  & 45.17 & 45.76 & 50.37 & 64.86 & 41.17 & 604 & 936 & 609 & 26083 & 140109 & 3115 & 6727 \\
        DiffMOT \citep{lv2024diffmot} [CVPR 2024]         & \textcolor{red}{52.90} & \textcolor{red}{49.70} & 55.23 & \textcolor{red}{73.13} & 44.36 & 754 & 781 & 614 & \textcolor{red}{15851} & 138622 & 2545 & \textcolor{blue}{6682} \\
        MOTIP \citep{gao2025motip} [CVPR 2025]            & 51.11 & 47.60 & \textcolor{blue}{55.48} & 67.13 & \textcolor{blue}{47.27} & \textcolor{blue}{787} & 884  & \textcolor{blue}{478} & 33595 & \textcolor{blue}{125932} & 4033 & 11350 \\
        TrackTrack \citep{shim2025TrackTrack} [CVPR 2025] & \textcolor{blue}{52.61} & \textcolor{blue}{49.62} & 54.92 & \textcolor{blue}{71.80} & 44.47 & 689 & 830  & 630 & \textcolor{blue}{18071} & 136878 & \textcolor{blue}{2309} & \textcolor{red}{5329} \\
        TCEI \citep{guo2026tcei} [CVPR 2026]              & 52.21 & 47.43 & \textcolor{red}{56.27} & 64.43 & \textcolor{red}{49.95} & \textcolor{red}{921} & 791  & \textcolor{red}{437} & 44604 & \textcolor{red}{114783} & 4706 & 10683 \\
        \bottomrule
    \end{tabular}
    \end{adjustbox}
\end{table}

\paragraph{Animal-group performance.}
YOLOX-X achieves 72.34 and 54.07 AP under Protocols~1 and~2 (See Table~\ref{tab:detection_overall}), so cross-protocol tracking results should be interpreted alongside detector performance. In Protocol~2, birds achieve 74.67 AP and 70.59 mean DiffMOT HOTA, versus 23.08 and 25.36 for other animals, revealing substantial variation across groups. Protocol~2 has no amphibian or reptile test categories (--). Appendix Tables~\ref{tab:detection_group_thresholds}, \ref{tab:detection_recall}, and \ref{tab:group_hota} provide further detection, recall, and tracking results.

\paragraph{Video-level difficulty.}
Within each protocol, we split the 315 test videos equally into Easy, Medium, and Hard groups by mean per-video HOTA across all eight baselines. DiffMOT leads all groups except Protocol~2 Hard, where TCEI and MOTIP achieve 29.08 and 28.27 HOTA, exceeding DiffMOT's 24.73 (Appendix Table~\ref{tab:difficulty_hota}). This ranking change highlights differences obscured by aggregate scores. Table~\ref{tab:video_hota_stats_full} and Figure~\ref{fig:video_hota_ecdf} in the appendix report the full distributions.

\begin{figure}[t]
    \centering
    \includegraphics[width=\linewidth]{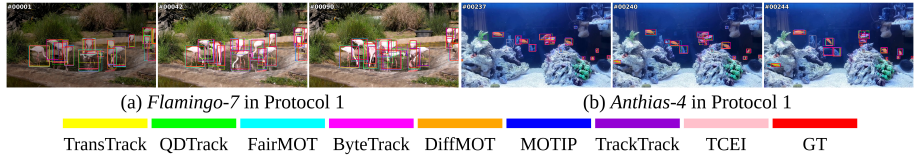}
    \caption{Qualitative comparison of eight trackers on Protocol~1.}
    \label{fig:qualitative_main}
\end{figure}

\paragraph{Qualitative comparison.}
Figure~\ref{fig:qualitative_main} compares eight trackers on Flamingo-7 and Anthias-4, illustrating localization differences under overlapping animals and pose changes, and incomplete coverage of small underwater targets. Appendix~\ref{app:qualitative} provides additional comparisons.

\subsection{CDA Ablation and Design Analysis}
\label{sec:exp_cda}

We analyze CDA with TrackTrack on Protocol~2, keeping cached detections, appearance features, track management, and post-processing identical. Original TrackTrack retains its scoring and gating, while DIoU, CDA, and the other geometric controls share normalized-DIoU gating at a threshold of 0.10 to compare association scores.

\begin{wraptable}{r}{0.55\textwidth}
    \centering
    \vspace{-0.1cm}
    \caption{Main CDA ablation on Protocol~2.}
    \label{tab:cda_main}
    \footnotesize
    \setlength{\aboverulesep}{0pt}
    \setlength{\belowrulesep}{0pt}
    \renewcommand{\arraystretch}{1.1}
    \begin{adjustbox}{width=\linewidth}
    \begin{tabular}{cccccc}
        \toprule
        \rowcolor{gray!12}
        & Configuration
        & HOTA
        & AssA
        & IDF1
        & IDSW$\downarrow$ \\
        \midrule
        \midrule
        \ablnum{1}
        & TrackTrack
        & 52.61 & 56.97 & 54.92 & 2309 \\
        \ablnum{2}
        & TrackTrack (DIoU)
        & 52.85 & 56.14 & 55.82 & 2133 \\
        \midrule
        \ablnum{3}
        & CDA ($\alpha=0$)
        & 53.83 & 58.29 & 57.03 & 2313 \\
        \ablnum{4}
        & CDA ($\alpha=0.25$)
        & 53.83 & 58.19 & 56.89 & 2054 \\
        \ablnum{5}
        & CDA ($\alpha=1$)
        & 53.56 & 58.12 & 56.28 & 2082 \\
        \rowcolor{VastMAT}
        \ablnum{6}
        & CDA-density (fwd)
        & \textbf{53.92}
        & \textbf{58.38}
        & \textbf{57.06}
        & 2166 \\
        \ablnum{7}
        & CDA-density (rev)
        & 53.54 & 57.85 & 56.26
        & \textbf{1973} \\
        \ablnum{8}
        & CDA ($\alpha=0.3824$)
        & 53.52 & 57.62 & 56.40 & 2048 \\
        \bottomrule
    \end{tabular}
    \end{adjustbox}
\end{wraptable}

\paragraph{Main CDA ablation on Protocol~2.}
In Table~\ref{tab:cda_main}, adaptive CDA (\ablnum{6}) improves HOTA from 52.61 to 53.92 over original TrackTrack (\ablnum{1}), with AssA and IDF1 gains of 1.41 and 2.14 percentage points. Compared with DIoU under identical gating (\ablnum{2}), HOTA and AssA improve by 1.07 and 2.24 points, while DetA changes only from 49.96 to 50.01, indicating that the gains primarily reflect better association. Pure center scoring (\ablnum{3}) and fixed fusion (\ablnum{4}) both achieve 53.83 HOTA, but incorporating IoU reduces IDSW from 2,313 to 2,054, showing that overlap information can reduce identity switches while preserving HOTA.

\paragraph{Normalization and geometric scores.}
To examine the scale reference for center distance, we fix $\alpha=0.5$ and vary only the normalization denominator. Mean-diagonal normalization achieves 53.80 HOTA, exceeding enclosing-box normalization at 53.18 and approaching minimum-diagonal normalization at 53.82 (See Table~\ref{tab:cda_denominator} in the appendix), supporting a distance reference based on the boxes' own scales. Comparisons with standard IoU variants further assess the complete geometric score\citep{rezatofighi2019generalized,zheng2020diou,zhang2022focal}. In Table~\ref{tab:cda_iou_variants}, CDA (\ablnum{5}) achieves the highest HOTA, AssA, and IDF1, whereas EIoU (\ablnum{4}) yields the fewest IDSW. These rankings show that identity-switch counts and overall identity consistency capture different aspects of performance, motivating evaluation with complementary association metrics.

\begin{wraptable}{r}{0.55\textwidth}
    \centering
    \vspace{-0.1cm}
    \caption{CDA versus IoU variants on Protocol~2.}
    \label{tab:cda_iou_variants}
    \footnotesize
    \setlength{\aboverulesep}{0pt}
    \setlength{\belowrulesep}{0pt}
    \renewcommand{\arraystretch}{1.1}
    \begin{adjustbox}{width=\linewidth}
    \begin{tabular}{cccccc}
        \toprule
        \rowcolor{gray!12}
        & Configuration
        & HOTA
        & AssA
        & IDF1
        & IDSW$\downarrow$ \\
        \midrule
        \midrule
        \ablnum{1}
        & TrackTrack
        & 52.61 & 56.97 & 54.92 & 2309 \\
        \ablnum{2}
        & GIoU
        & 53.50 & 57.51 & 56.56 & 2012 \\
        \ablnum{3}
        & CIoU
        & 52.96 & 56.40 & 56.00 & 2115 \\
        \ablnum{4}
        & EIoU
        & 53.61 & 57.97 & 56.39
        & \textbf{1960} \\
        \rowcolor{VastMAT}
        \ablnum{5}
        & CDA ($\alpha=0.25$)
        & \textbf{53.83}
        & \textbf{58.19}
        & \textbf{56.89}
        & 2054 \\
        \bottomrule
    \end{tabular}
    \end{adjustbox}
\end{wraptable}

\paragraph{Adaptive fusion.}
We compare fixed weights with forward and reverse density rules to assess the effect of adapting weights to detection count. In Table~\ref{tab:cda_main}, the forward rule, fwd (\ablnum{6}), achieves 53.92 HOTA, exceeding 53.83 for fixed $\alpha=0.25$ (\ablnum{4}), the reverse rule, rev (\ablnum{7}), and the fixed reference $\alpha=0.3824$ averaged over diagnostic events (\ablnum{8}). The forward--reverse difference shows that the direction of weight adaptation matters, emphasizing IoU at higher detection counts and center similarity at lower counts. Nine density-parameter configurations outperform TrackTrack (See Table~\ref{tab:cda_density_grid} in the appendix). Appendix~\ref{app:cda_details} provides complete weight, post-processing, and candidate-pair diagnostics.
\vspace{-5pt}
\subsection{CDA Across Trackers and Protocols}
\label{sec:cross_tracker}

\noindent\textbf{Cross-tracker results.}
With fixed $\alpha=0.25$, CDA improves HOTA by 1.22 and 1.36 percentage points for TrackTrack (\ablnum{1} and \ablnum{2}) and ByteTrack (\ablnum{3} and \ablnum{4}), respectively (See Table~\ref{tab:cda_cross_tracker}). Their IDF1 scores increase from 54.92 and 50.37 to 56.89 and 52.96, accompanied by higher AssA and fewer IDSW. DiffMOT (\ablnum{5} and \ablnum{6}) gains slightly in HOTA and IDF1 without corresponding improvements in AssA or IDSW; its weight sensitivity is analyzed below.

\begin{wraptable}{r}{0.55\textwidth}
    \centering
    \vspace{-0.1cm}
    \caption{Cross-tracker results on Protocol~2.}
    \label{tab:cda_cross_tracker}
    \footnotesize
    \setlength{\aboverulesep}{0pt}
    \setlength{\belowrulesep}{0pt}
    \renewcommand{\arraystretch}{1.1}
    \begin{adjustbox}{width=\linewidth}
    \begin{tabular}{cccccc}
        \toprule
        \rowcolor{gray!12}
        & Configuration
        & HOTA
        & AssA
        & IDF1
        & IDSW$\downarrow$ \\
        \midrule
        \midrule
        \ablnum{1}
        & TrackTrack
        & 52.61 & 56.97 & 54.92 & 2309 \\
        \rowcolor{VastMAT}
        \ablnum{2}
        & TrackTrack + CDA
        & \textbf{53.83}
        & \textbf{58.19}
        & \textbf{56.89}
        & \textbf{2054} \\
        \midrule
        \ablnum{3}
        & ByteTrack
        & 45.17 & 48.55 & 50.37 & 3115 \\
        \rowcolor{VastMAT}
        \ablnum{4}
        & ByteTrack + CDA
        & \textbf{46.53}
        & \textbf{51.27}
        & \textbf{52.96}
        & \textbf{2292} \\
        \midrule
        \ablnum{5}
        & DiffMOT
        & 52.90
        & \textbf{57.61}
        & 55.23
        & \textbf{2545} \\
        \rowcolor{VastMAT}
        \ablnum{6}
        & DiffMOT + CDA
        & \textbf{52.93}
        & 57.30
        & \textbf{55.54}
        & 2711 \\
        \bottomrule
    \end{tabular}
    \end{adjustbox}
\end{wraptable}

\noindent\textbf{Candidate acceptance in ByteTrack.}
ByteTrack's geometric cost $1-\mathrm{IoU}$ and matching threshold of 0.6 imply $\mathrm{IoU}\geq0.4$. With CDA, this becomes $S\geq0.4$, admitting some low-overlap candidates. FN and IDSW decrease by 2,839 and 823, respectively, while FP increases by 6,743 and MOTA falls by 0.99 percentage points. Thus, the change in candidate acceptance improves object coverage and identity association but also introduces more false positives.

\noindent\textbf{Tracker-specific fusion weights.}
Figure~\ref{fig:alpha_curves} shows that TrackTrack achieves higher HOTA when center similarity receives greater weight. Among the tested weights, DiffMOT performs best at $\alpha=0.5$, reaching 53.53 HOTA, a gain of 0.63 percentage points (See Table~\ref{tab:diffmot_alpha} in the appendix). These results indicate that suitable fusion weights depend on the base tracker.

\noindent\textbf{Results on Seen categories.}
On Protocol~1, fixed and adaptive CDA improve TrackTrack's HOTA from 65.10 to 66.66 and 66.68, respectively; the adaptive variant also reduces IDSW from 1,962 to 1,652. Together with Protocol~2, these results show that CDA improves TrackTrack on both Seen and Unseen categories. Table~\ref{tab:cda_protocol1} in the appendix reports the full metrics.

\noindent\textbf{Inference cost.}
With detections and appearance features cached, tracking 93,111 frames takes 129.32s compared with 128.97s for the baseline, an increase of only 0.27\%. Table~\ref{tab:inference_cost} in the appendix provides a similarity-computation microbenchmark.

\section{Conclusion}
\label{sec:conclusion}
\vspace{-5pt}
We introduce VastMAT, comprising 337 animal categories and 2947 videos with continuous multi-instance identity annotations and Seen- and Unseen-category evaluation protocols. Evaluating eight MOT methods reveals performance variation across animal groups and videos. We further propose CDA, which requires no additional training and improves TrackTrack's HOTA by 1.58 and 1.31 percentage points under the two protocols. VastMAT provides a unified benchmark for detection, identity association, and Cross-category generalization across diverse animals.


\section*{Limitations}
\label{sec:limitations}
VastMAT inherits category imbalance and scene biases from publicly available online videos, so benchmark results may not fully reflect performance on underrepresented animals and environments. Moreover, CDA relies on the base tracker's detection candidates and motion predictions, and its gains vary across trackers and fusion settings.

\subsection*{AI Use Statement}
We used AI tools to assist with writing, language refinement, related-work discovery, and drafting portions of the manuscript. All AI-assisted content was reviewed and verified by the authors, who take full responsibility for the final paper.

\subsection*{Ethics Statement}
VastMAT is constructed from existing public videos, whose rights remain with their respective holders.
Release materials will comply with individually verified licenses or permissions and specify separate terms for code, annotations, and third-party videos; the dataset license does not extend rights to third-party content. The project will provide a channel for copyright and privacy concerns. Verified issues will lead to access restrictions, corrections, or removal, with effects on the data and evaluation scope documented.

\subsection*{Reproducibility Statement}
At release, we will provide the category dictionary, split lists for both protocols and the training-side development set, annotation format and data preparation tools, CDA implementation, and unified evaluation code. We will also release baseline training and inference configurations, pretraining sources, predictions, and metric-generation scripts. Annotation-quality materials will include audit lists, independent reannotations, and scripts for box-level and within-clip identity consistency. Each result will be tied to explicit data, annotation, and evaluation versions. Appendix~\ref{app:maintenance_ethics} describes media access policies.

\bibliography{references}
\bibliographystyle{iclr2027_conference}

\clearpage
\setlength{\intextsep}{4pt plus 2pt minus 2pt}
\setlength{\textfloatsep}{4pt plus 2pt minus 2pt}
\setlength{\abovecaptionskip}{4pt}
\setlength{\belowcaptionskip}{0pt}

\appendix

\newcommand{\appendixfigure}[2]{%
  \IfFileExists{#1}{%
    \includegraphics[width=\linewidth,height=#2,keepaspectratio]{#1}%
  }{%
    \figureplaceholder{#1}{#2}%
  }%
}

\section*{Appendix}

To better understand VastMAT and CDA in this work, we provides additional visual results, dataset documentation, annotation validation, and experimental analyses. It is organized as follows:
\begin{itemize}
    \setlength{\itemsep}{4pt}
    \setlength{\parsep}{2pt}
    \setlength{\parskip}{2pt}
    \item \textbf{~\ref{app:qualitative}  Qualitative Results}
    
    Visual comparisons of eight baseline trackers across animal scenes.
    \item \textbf{~\ref{app:dataset_statistics}  Dataset Categories and Statistics}
    
    The category inventory, video distributions, co-occurrence, and visual similarity.
    \item \textbf{~\ref{app:annotation_details}  Annotation Protocol and Quality Assessment}
    
    The annotation format, review procedure, and independent reannotation audit.
    \item \textbf{~\ref{app:benchmark_results}          Additional Benchmark Results}
    
    Detection, animal-group tracking, and video-level performance analyses.
    \item \textbf{~\ref{app:cda_details}  CDA Implementation and Additional Experiments}
    
    Association settings, design comparisons, sensitivity analyses, diagnostics, and runtime.
    \item \textbf{~\ref{app:maintenance_ethics}  Dataset Documentation and Maintenance}
    
    Data access, versioning, issue reporting, and intended use.
\end{itemize}

\section{Qualitative Results}
\label{app:qualitative}

\begin{figure}[H]
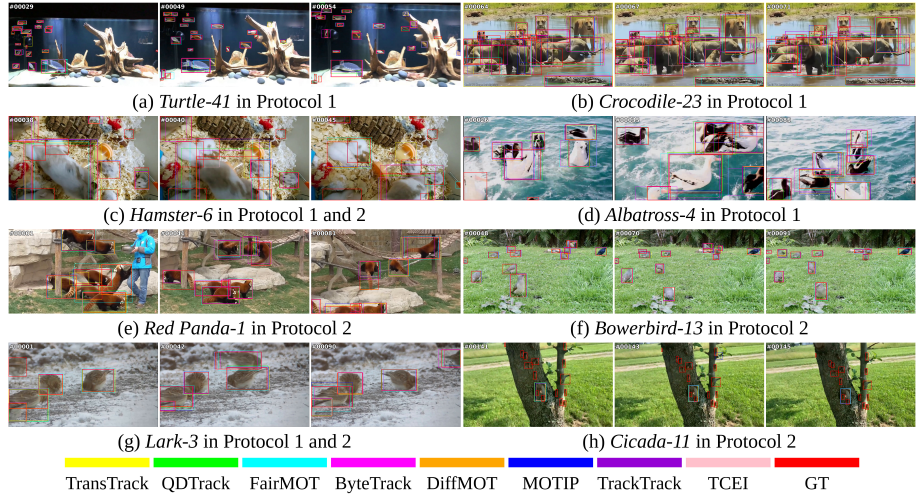

    \centering
\centering
    \appendixfigure{Figures/Fig_Qualitative.pdf}{0.52\textheight}
    \caption{
        Qualitative comparison of eight baseline trackers on selected VastMAT sequences.
    }
    \label{fig:qualitative}
\end{figure}

The main text presents representative examples in Figure~\ref{fig:qualitative_main}; this appendix provides more detailed qualitative comparisons in Figure~\ref{fig:qualitative}.

Figure~\ref{fig:qualitative} compares eight baseline trackers with ground truth across diverse animal scenes. Each triplet shows three sampled frames from one sequence, with frame indices in the upper-left corners. Colors identify trackers according to the legend; red boxes denote ground truth. Hamster-6 illustrates motion blur and variation in predicted box extent, while Crocodile-23 and Red Panda-1 contain overlapping animals and partial occlusion. Bowerbird-13 and Cicada-11 show incomplete prediction coverage for some small or low-contrast targets. The underwater scene in Turtle-41 and wave clutter in Albatross-4 further illustrate the diversity of imaging conditions. Together, these examples highlight scene-specific challenges in object coverage and localization that complement the quantitative evaluation.In Figure~\ref{fig:qualitative}, panels (c) and (g) are generated using Protocol~1 boxes, and "Protocol~1 and~2" in the caption means that the video appears in the test sets of both protocols.

\section{Dataset Categories and Statistics}
\label{app:dataset_statistics}

This section documents the category inventory and provides additional statistics on video composition, category co-occurrence, and visual similarity.

\subsection{Animal Categories and Grouping}
\label{app:animal_grouping}

\begin{wraptable}{r}{0.44\textwidth}
    \centering
    \footnotesize
    \vspace{-0.3cm}
    \caption{Animal groups in VastMAT and their Protocol~2 category counts.}
    \label{tab:animal_group_coverage}
    \setlength{\aboverulesep}{0pt}
    \setlength{\belowrulesep}{0pt}
    \renewcommand{\arraystretch}{1.1}
    \begin{adjustbox}{width=\linewidth}
    \begin{tabular}{lccc}
        \toprule
        \rowcolor{gray!12}
        Group & All & P2 Seen & P2 Unseen \\
        \midrule
        Mammals    & 82  & 65  & 17 \\
        Birds      & 85  & 74  & 11 \\
        Fish       & 117 & 103 & 14 \\
        Amphibians & 4   & 4   & 0  \\
        Reptiles   & 7   & 7   & 0  \\
        Others     & 42  & 30  & 12 \\
        \midrule
        Total      & 337 & 283 & 54 \\
        \bottomrule
    \end{tabular}
\end{adjustbox}
\end{wraptable}

VastMAT contains 337 animal categories. To summarize coverage and support group-wise evaluation, we assign them to six coarse groups: mammals, birds, fish, amphibians, reptiles, and other animals. Group membership follows the animal category rather than the filming environment. For example, Dolphin, Killer Whale, and Whale are mammals; Penguin is a bird; and Shark and Ray are fish. Other animals include the insects, arachnids, crustaceans, mollusks, and other invertebrates represented in the dataset.

Table~\ref{tab:animal_group_coverage} summarizes the six groups and their Protocol~2 training and test category counts. The complete category names are listed below; the accompanying category dictionary defines the mapping from category IDs to English names.

\begingroup
\small

\paragraph{Mammals (82 categories).}
Alpaca, Anteater, Antelope, Ape, Armadillo, Babirusa, Baboon,
Badger, Bat, Bear, Beaver, Bison, Buffalo, Camel, Cat, Cattle,
Cheetah, Chimpanzee, Civet, Colugo, Deer, Dog, Dolphin, Donkey,
Echidna, Elephant, Fox, Giraffe, Goat, Gorilla, Guinea Pig,
Hamster, Hedgehog, Hippopotamus, Horse, Hyena, Kangaroo,
Killer Whale, Koala, Lemur, Leopard, Lion, Lynx, Manatee,
Marmot, Mongoose, Monkey, Mouse, Orangutan, Otter, Panda,
Pangolin, Pig, Pika, Platypus, Porcupine, Porpoise, Possum,
Quokka, Rabbit, Racoon, Red Panda, Rhinoceros, Sea Lion,
Seal, Sheep, Sloth, Slow Loris, Snow Leopard, Squirrel,
Stoat, Tapir, Tiger, Treeshrew, Walrus, Whale, Wildebeest,
Wolf, Wolverine, Wombat, Yak, Zebra.

\paragraph{Birds (85 categories).}
Albatross, Bittern, Bluethroat, Bowerbird, Bulbul, Bustard,
Buzzard, Chicken, Cormorant, Cowbird, Crane, Crow, Cuckoo,
Curassow, Dipper, Drongo, Duck, Eagle, Falcon, Finch,
Flamingo, Frigatebird, Goldcrest, Goldeneye, Goose,
Great Argus, Grebe, Gull, Harrier, Heron, Hoopoe, Hornbill,
Hummingbird, Ibis, Kingfisher, Kiwi, Lapwing, Lark, Myna,
Nightingale, Nuthatch, Oriole, Ostrich, Owl, Parrot,
Peacock, Pelican, Penguin, Pigeon, Pipit, Plover, Puffin,
Quail, Rail, Robin, Sandpiper, Secretarybird, Shearwater,
Shoebill, Shorebird, Shrike, Sparrow, Starling, Stilt,
Stork, Swallow, Swan, Tern, Thick-knee, Thrush, Tinamou,
Tit, Toucan, Turkey, Vulture, Wagtail, Warbler, Waterhen,
Woodpecker, Wren, Wryneck, Jays, Cardinal bird, Magpie,
Blackbird.

\paragraph{Fish (117 categories).}
Angelfish, Anglerfish, Anthias, Arapaima, Archerfish,
Arowana, Bala Shark, Barb, Barracuda, Bass, Batfish,
Betta, Bichir, Billfish, Blind cavefish, Burbot,
Butterflyfish, Cardinalfish, Carp, Catfish, Chimaera,
Cichlid, Cleaner Wrasse, Climbing perch, Clownfish, Cod,
Coelacanth, Boxfish, Damselfish, Danio, Datnoid, Discus,
Dottyback, Drum fish, Eel, Electric eel, Elephantnose fish,
Fairy Wrasse, Fangblenny, Flounder, Flowerhorn, Flying fish,
Fusilier, Gar, Glassy fish, Goby, Goldfish, Gourami,
Grayling, Grouper, Guppy, Hatchetfish, Hawkfish, Herring,
Hillstream loach, Hogfish, Knifefish, Lionfish, Loach,
Lungfish, Mandarinfish, Milkfish, Molly, Moorish Idol,
Mudskipper, Needlefish, Ocean Sunfish, Oscar, Paddlefish,
Paradise fish, Parrotfish, Payara, Perch, Pike,
Pineapplefish, Piranha, Pufferfish, Rabbitfish,
Rainbowfish, Ray, Rockfish, Royal Gramma, Sailfin Tang,
Salmon, Scorpionfish, Sculpin, Seahorse, Shark,
Silver Dollar, Snailfish, Snakehead, Snapper, Snipefish,
Squirrelfish, Sturgeon, Sunfish, Sweetlips, Taimen, Tangs,
Tench, Tetra, Tilapia, Toadfish, Trevally, Triggerfish,
Trout, Trumpetfish, Tuna, Unicornfish,
White cloud mountain minnow, Wolffish, Wrasse,
Yellow Wrasse, Zander, Mackerel, Sixline Wrasse, Tarpon.

\paragraph{Amphibians (4 categories).}
Frog, Giant Salamander, Newt, Toad.

\paragraph{Reptiles (7 categories).}
Chameleon, Crocodile, Gecko, Lizard, Snake, Tortoise, Turtle.

\paragraph{Other animals (42 categories).}
Ant, Bee, Beetle, Brittle Star, Butterfly, Caterpillar,
Centipede, Cicada, Cockroach, Crab, Cricket, Cuttlefish,
Dragonfly, Earthworm, Feather Star, Firefly, Fly,
Grasshopper, Horseshoe Crab, Jellyfish, Ladybug,
Leaf Insect, Lobster, Mantis, Mealworm, Millipede,
Mosquito, Moth, Nautilus, Octopus, Scallop, Scorpion,
Sea Anemone, Sea Spider, Shrimp, Snail, Spider, Squid,
Starfish, Stick Insect, Termite, Wasp.

\endgroup

\subsection{Video-Level Distributions}
\label{app:video_distributions}

\begin{figure}[htbp]
    \centering
\centering
    \appendixfigure{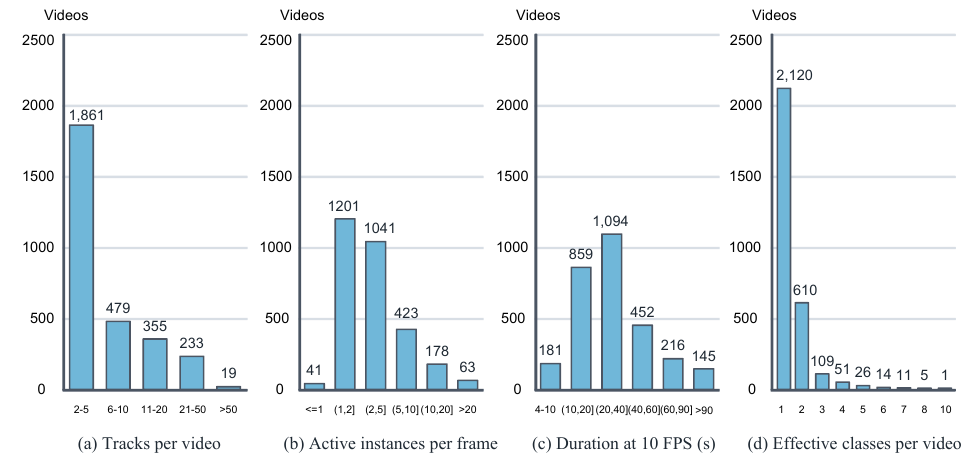}{0.40\textheight}
    \caption{Video-level distributions of (a) trajectory count, (b) average active instances per frame, (c) duration, and (d) category count in VastMAT.}
    \label{Figsiweiduibi}
\end{figure}

Figure~\ref{Figsiweiduibi} summarizes four video-level attributes: \textbf{trajectory count}, \textbf{average active instances per frame}, \textbf{video duration}, and \textbf{valid category count}. In Figure~\ref{Figsiweiduibi}(a), 607 videos contain more than 10 trajectories, including 252 with at least 21, with a maximum of 152 per video. Figure~\ref{Figsiweiduibi}(b) shows that 664 videos average more than 5 active instances per frame, including 63 with more than 20. In Figure~\ref{Figsiweiduibi}(c), over half the videos last 20--60 seconds, while 145 exceed 90 seconds. Figure~\ref{Figsiweiduibi}(d) shows 2,120 single-category and 827 multi-category videos, with up to 10 categories per video. VastMAT thus covers sparse scenes, dense groups, long sequences, and multi-category interactions.

\subsection{Category Coverage and Co-occurrence Structure}
\label{app:category_coverage}

\begin{figure}[htbp]
    \centering
\centering
    \appendixfigure{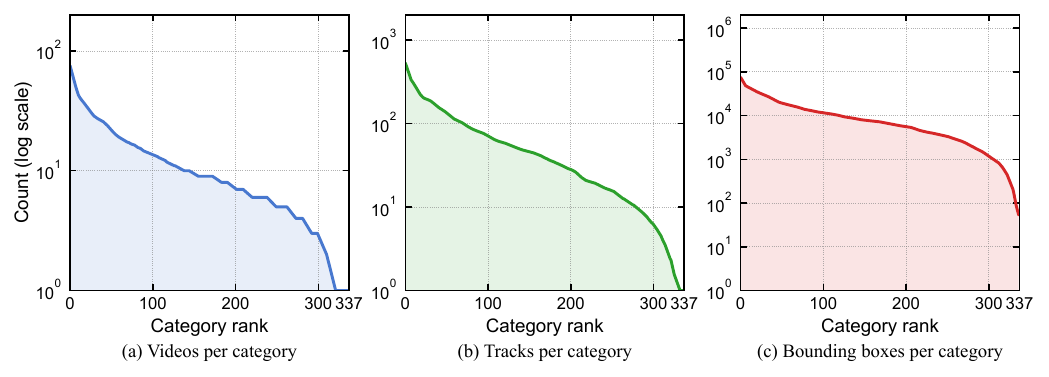}{0.5\textheight}
    \caption{Per-category distributions of videos, trajectories, and bounding boxes in VastMAT.}
    \label{Fig_datasetchangwei}
\end{figure}

\paragraph{Per-category distributions.}
Figure~\ref{Fig_datasetchangwei} shows per-category distributions of videos, trajectories, and bounding boxes.

\begin{figure}[H]
    \centering
    \vspace{-1cm}
    \appendixfigure{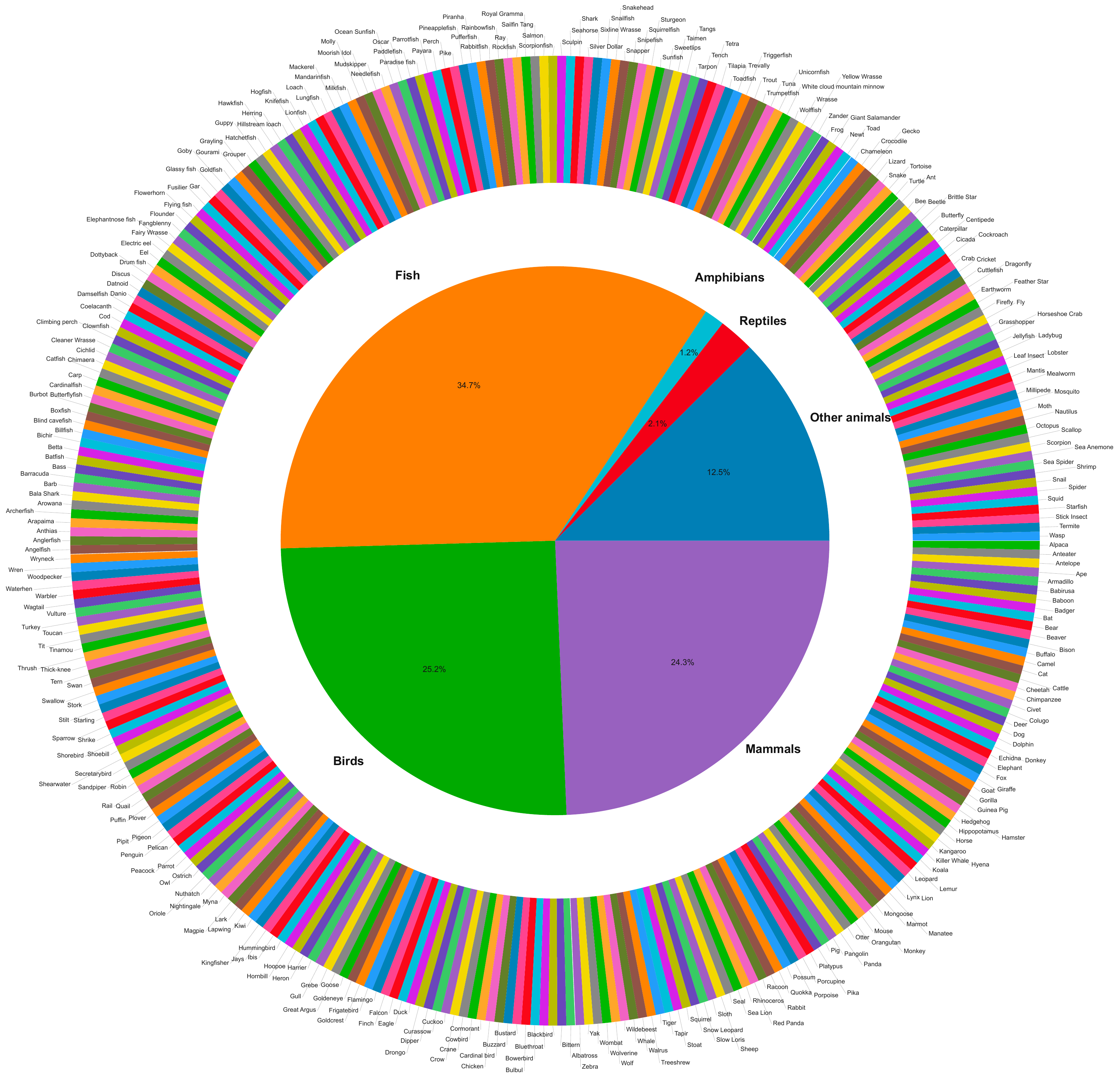}{0.82\textheight}
    \caption{All 337 categories in VastMAT grouped into six major animal groups, with inner-ring proportions and outer-ring category labels.}
    \label{Fig337tongji}
\end{figure}

\paragraph{Complete category coverage.}
Figure~\ref{Fig337tongji} visualizes the distribution of all 337 categories across six major animal groups. The inner donut chart reports the category counts and proportions: Fish (117 categories, 34.7\%), Birds (85, 25.2\%), Mammals (82, 24.3\%), Other animals (42, 12.5\%), Amphibians (4, 1.2\%), and Reptiles (7, 2.1\%). The outer ring further breaks down each major group into its constituent categories, showing complete coverage of the 337 categories. 

\paragraph{Frequent categories and co-occurrence.}

\begin{figure}[H]
    \centering
    \appendixfigure{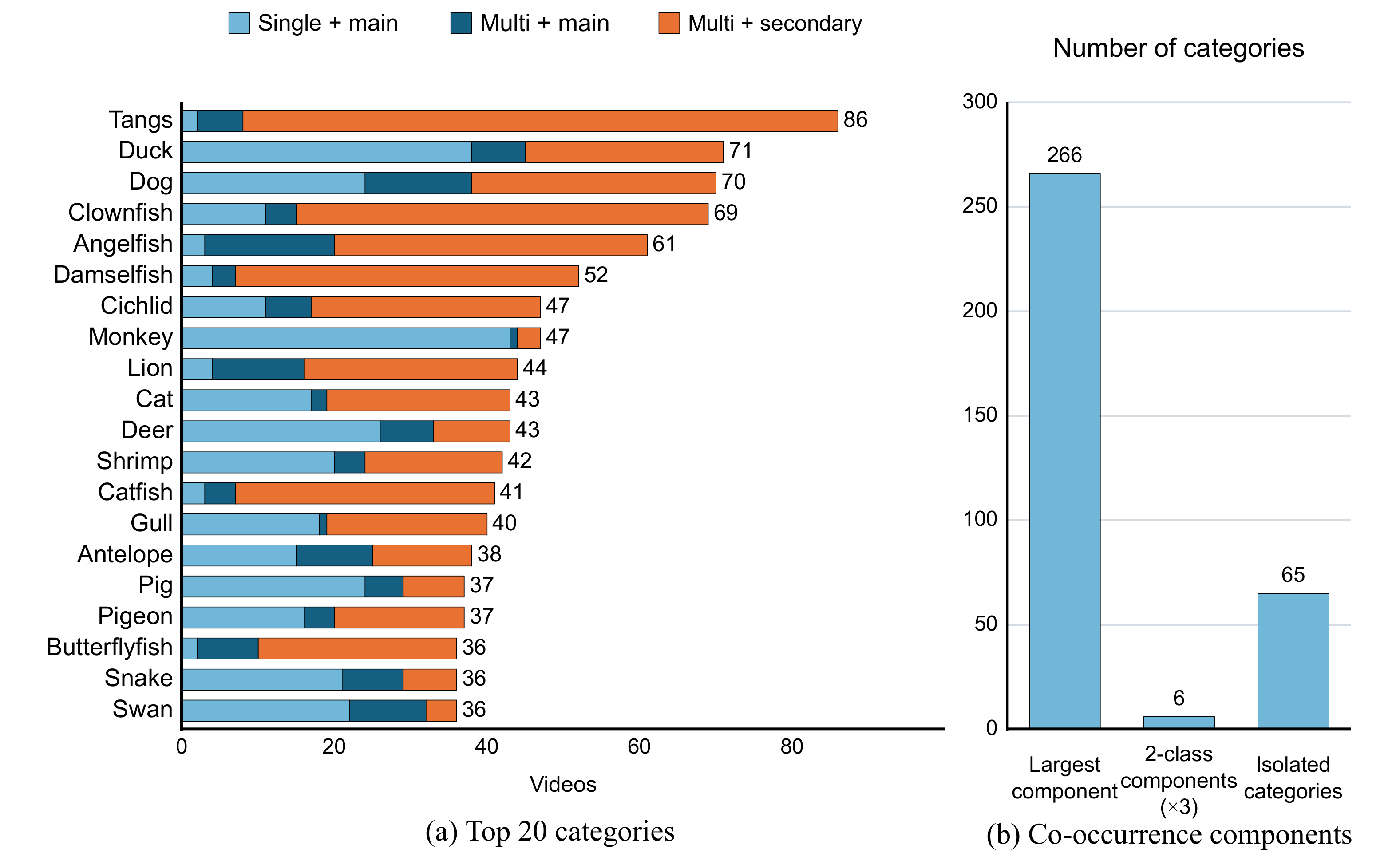}{0.74\textheight}
    \caption{Category statistics in VastMAT: (a) composition of the 20 most frequent categories and (b) connected components in the category co-occurrence graph.}
    \label{Figtop20}
\end{figure}

Figure~\ref{Figtop20}(a) shows distinct scene compositions across categories. Tangs, Clownfish, and Damselfish often appear as secondary categories in multi-category videos, whereas Duck and Monkey more often appear as the primary subject. Multi-category videos therefore increase scene complexity while broadening coverage of naturally co-occurring categories.

Figure~\ref{Figtop20}(b) shows the connected components of the category co-occurrence graph. The largest contains 266 categories, with 3 additional two-category components and 65 isolated categories, indicating that most categories are linked through multi-category videos. This structure requires category-disjoint splits to assign entire connected components, preventing categories from the same video from entering both training and testing.

\subsection{Visual Similarity Across Categories}

\label{app:visual_similarity}

\begin{figure}[H]
    \centering
\centering
    \appendixfigure{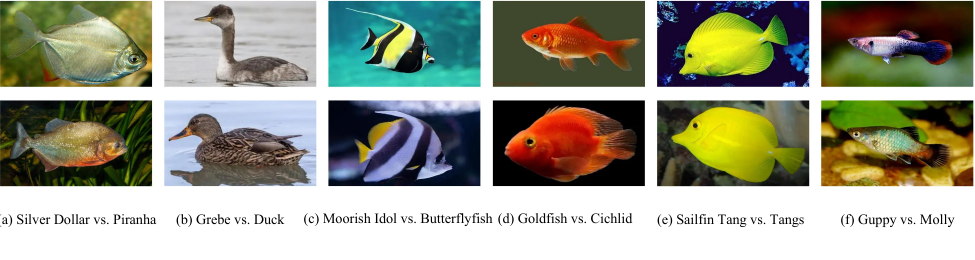}{0.65\textheight}
    \caption{Representative visually similar category pairs in VastMAT: (a) Silver Dollar vs. Piranha, (b) Grebe vs. Duck, (c) Moorish Idol vs. Butterflyfish, (d) Goldfish vs. Cichlid (Parrot Cichlid), (e) Yellow Sailfin Tang vs. Tangs (Yellow Tang), and (f) Guppy vs. Molly.}

    \label{Figxilidu}
\end{figure}

VastMAT includes visually similar animals from different categories among its 337 categories. Figure~\ref{Figxilidu} shows six representative pairs: Silver Dollar and Piranha, and Guppy and Molly, have similar body outlines; Grebe and Duck share head, neck, and torso structures in side views; Moorish Idol and Butterflyfish have similar stripes and body shapes; and Goldfish and Cichlid, and Yellow Sailfin Tang and Tangs, closely resemble one another in color and overall appearance.

Differences between these categories often lie in local shape and texture, which low resolution, motion blur, pose changes, and occlusion can obscure. In multi-category videos, similar-looking individuals may yield similar association features, complicating cross-frame identity discrimination. These examples illustrate instance-level appearance ambiguity in the benchmark.

\section{Annotation Protocol and Quality Assessment}
\label{app:annotation_details}

This section specifies the annotation interface and documents the inspection procedure and independent quality audit.

\subsection{Annotation Format}
\label{app:annotation_format}

Annotations for each video are stored in a nine-column gt.txt file, with fields defined in Table~\ref{tab:gt_format}.

\begin{table}[htbp]
    \centering
    \caption{The nine-column annotation interface of \texttt{gt.txt} in VastMAT.}
    \label{tab:gt_format}
    \centering
    \footnotesize
    \setlength{\tabcolsep}{4pt}
    \setlength{\aboverulesep}{0pt}
    \setlength{\belowrulesep}{0pt}
    \renewcommand{\arraystretch}{1.1}
    \begin{adjustbox}{width=\linewidth}
    \begin{tabularx}{\linewidth}{c l L}
        \toprule
        \rowcolor{gray!12}
        Position & Name & Description \\
        \midrule
        1 & Frame number & Index of the annotated frame; starts at 1. \\
        2 & Identifier & Video-local unique track ID; IDs need not be consecutive. \\
        3 & Box left & Left $x$-coordinate of the axis-aligned bounding box, in pixels. \\
        4 & Box top & Top $y$-coordinate of the axis-aligned bounding box, in pixels. \\
        5 & Box width & Width of the axis-aligned bounding box, in pixels. \\
        6 & Box height & Height of the axis-aligned bounding box, in pixels. \\
        7 & Confidence
        & Annotation flag; set to 1 for valid annotated targets. \\
        8 & Class & Global class ID mapped by the released category dictionary. \\
        9 & Visibility
        & Reserved visibility field; set to $-1$ when unspecified. This value does not mark an ignored target. \\
        \bottomrule
    \end{tabularx}
    \end{adjustbox}
\end{table}

\subsection{Annotation Inspection and Quality Assessment}

\label{app:annotation_quality}

\paragraph{Manual inspection and correction.}
Annotation review examines object coverage, box localization, category labels, and identity continuity through continuous playback and frame-by-frame inspection. Playback reveals identity correspondences during entry, exit, occlusion, and interactions, while individual frames are checked for missed objects, duplicate annotations, and box placement. Corrections cover the affected frame and relevant neighboring frames. Propagation drift is checked along subsequent frames, and identity errors are traced along the associated trajectories.

\paragraph{Sampling and independent reannotation.}
We select 48 videos through stratified random sampling across all six animal groups, with the sample composition reported in Table~\ref{tab:annotation_audit_groups}.
Two non-overlapping, consecutive 20-frame clips are randomly selected from each video,
yielding 96 clips and 1,920 frames. Reannotators receive only images without annotation overlays and the shared guidelines, without access to the original annotation files. Reannotation follows the same rules for visible-region boxes, occlusion handling, category mapping, and within-clip identity preservation. 9 frames contain no boxes in either annotation set; the remaining 1,911 contain annotations from at least one set.

\paragraph{Agreement metrics.}
We use the Hungarian algorithm for one-to-one box matching within each frame,
accepting matches with $\mathrm{IoU}\geq0.5$.
Let $N_{\mathrm{o}}$, $N_{\mathrm{r}}$, and $M$ denote the original box count, reannotated box count, and valid matched-pair count, respectively.
The two matching coverage rates and box-matching F1 are defined as
\begin{equation}
C_{\mathrm{o}}=\frac{M}{N_{\mathrm{o}}},
\qquad
C_{\mathrm{r}}=\frac{M}{N_{\mathrm{r}}},
\qquad
F_1=\frac{2M}{N_{\mathrm{o}}+N_{\mathrm{r}}}.
\label{eq:annotation_agreement}
\end{equation}
Mean IoU is computed only over valid matched pairs, and category agreement compares their labels. These metrics quantify correspondence between the annotation sets, while unmatched boxes identify disagreements in object coverage or localization.

\begin{wraptable}{r}{0.6\textwidth}
    \centering
    \vspace{-0.3cm}
    \caption{Independent reannotation audit by animal group. Coverage and F1 are percentages.}
    \label{tab:annotation_audit_groups}
    \footnotesize
    \setlength{\tabcolsep}{4pt}
    \setlength{\aboverulesep}{0pt}
    \setlength{\belowrulesep}{0pt}
    \renewcommand{\arraystretch}{1.1}
    \begin{adjustbox}{width=\linewidth}
    \begin{tabular}{lccccccc}
        \toprule
        \rowcolor{gray!12}
        Group & Videos & $N_{\mathrm{o}}$ & $N_{\mathrm{r}}$
        & $M$ & $C_{\mathrm{o}}$ & $C_{\mathrm{r}}$ & $F_1$ \\
        \midrule
        Mammals    & 8 & 1,093 & 1,116 & 1,079 & 98.72 & 96.68 & 97.69 \\
        Birds      & 8 & 1,029 & 1,032 &   996 & 96.79 & 96.51 & 96.65 \\
        Fish       & 8 & 1,948 & 1,941 & 1,878 & 96.41 & 96.75 & 96.58 \\
        Amphibians & 7 &   914 &   926 &   884 & 96.72 & 95.46 & 96.09 \\
        Reptiles   & 8 &   601 &   715 &   575 & 95.67 & 80.42 & 87.39 \\
        Others     & 9 & 2,300 & 2,280 & 2,220 & 96.52 & 97.37 & 96.94 \\
        \midrule
        Overall    & 48 & 7,885 & 8,010 & 7,632 & 96.79 & 95.28 & 96.03 \\
        \bottomrule
    \end{tabular}
\end{adjustbox}
\end{wraptable}

\paragraph{Overall and group-wise results.}
The audit contains 7,885 original and 8,010 independently reannotated boxes, with 7,632 valid matches. Matching coverage is $96.79\%$ for the original annotations and $95.28\%$ for reannotations; box-matching F1 is $96.03\%$, and mean matched-box IoU is $0.9555$. All 7,632 matched pairs agree in category. The original and reannotated sets contain 253 and 378 unmatched boxes, respectively.

\begin{wraptable}{r}{0.6\textwidth}
    \centering
    \vspace{-0.2cm}
    \caption{Original-side matching coverage by object size.}
    \label{tab:annotation_audit_size}
    \footnotesize
    \setlength{\aboverulesep}{0pt}
    \setlength{\belowrulesep}{0pt}
    \renewcommand{\arraystretch}{1.1}
    \begin{adjustbox}{width=\linewidth}
    \begin{tabular}{lccc}
        \toprule
        \rowcolor{gray!12}
        Box area $a$ (pixels$^2$)
        & Original boxes & Unmatched & $C_{\mathrm{o}}$ (\%) \\
        \midrule
        $a < 32^2$                  &   310 & 59 & 80.97 \\
        $32^2 \leq a < 64^2$         & 1,120 & 60 & 94.64 \\
        $64^2 \leq a < 128^2$        & 1,508 & 66 & 95.62 \\
        $a \geq 128^2$              & 4,947 & 68 & 98.63 \\
        \bottomrule
    \end{tabular}
\end{adjustbox}
\end{wraptable}

Table~\ref{tab:annotation_audit_groups} reports group-wise results. Matching coverage exceeds $95\%$ on both sides for all five groups except reptiles. Of the 140 unmatched reannotated reptile boxes, 134 come from Snake-5, Snake-61, and Lizard-1, indicating that disagreement is concentrated in a few sequences. Overall metrics aggregate all audited boxes and describe agreement within this sample.

\paragraph{Object-scale analysis.}
We group original boxes by their pixel area in the audited images (See Table~\ref{tab:annotation_audit_size}).
Matching coverage on the original side increases with object size,
from $80.97\%$ for the smallest group to $98.63\%$ for the largest.
Objects with area below $32^2$ pixels account for $3.93\%$ of original boxes
but $23.32\%$ of unmatched original boxes,
indicating that small objects warrant particular attention during annotation review.

\paragraph{Within-clip identity consistency.}
We assess temporal identity correspondence by mapping original identities to reannotated identities within each 20-frame clip. An original identity matched in at least two frames is consistent if every matched frame maps to the same reannotated identity. Within-clip identity consistency is the fraction of eligible clip--identity pairs satisfying this condition; the same identity in different audited clips is counted separately. We also report majority coverage and the longest consecutive consistent span to characterize the stability of within-clip correspondence.

\begin{wraptable}{r}{0.6\textwidth}
    \centering
    \footnotesize
    \vspace{-0.2cm}
    \caption{Within-clip identity consistency in the independent reannotation audit.}
    \label{tab:annotation_audit_id}
    \setlength{\aboverulesep}{0pt}
    \setlength{\belowrulesep}{0pt}
    \renewcommand{\arraystretch}{1.1}
    \begin{adjustbox}{width=\linewidth}
    \begin{tabular}{lc}
        \toprule
        \rowcolor{gray!12}
        Metric & Value \\
        \midrule
        Analyzed clips & 96 \\
        Eligible clip--identity pairs (matched in $\geq2$ frames) & 427 \\
        Pairs mapped to a single reannotated identity & 424 \\
        Within-clip identity consistency & 99.30\% \\
        Mean majority coverage & 99.81\% \\
        Mean longest consecutive consistent span & 18.49 frames \\
        Videos with consistent identities in all audited clips & 46 / 48 \\
        \bottomrule
    \end{tabular}
    \end{adjustbox}
\end{wraptable}

The 96 audited clips contain 427 eligible clip--identity pairs, of which 424 map to a single reannotated identity, yielding $99.30\%$ within-clip identity consistency. Mean majority coverage is $99.81\%$, and the mean longest consecutive consistent span is $18.49$ frames. Among the 48 videos, 46 have a within-clip IDR of $1.00$, while 2 fall below $1.00$: the second clip of Crocodile-2 has an IDR of $0.50$ (only 2 GT IDs), and the second clip of Bee-7 has an IDR of $0.93$ (2 of 30 densely distributed GT IDs change identity). Both involve brief within-clip identity assignment changes in scenes with changing object counts or dense small objects,
consistent with entry and exit under occlusion and candidate competition. Table~\ref{tab:annotation_audit_id} summarizes the clip-level identity results.

This metric evaluates 20-frame clips and complements box-level agreement; identity re-entry after prolonged occlusion can be further assessed on full trajectories.

\section{Additional Benchmark Results}
\label{app:benchmark_results}

This section supplements Section~\ref{sec:evaluation_results} with detection results and tracking analyses by animal group and video difficulty.

\subsection{Additional Detection Results}

\label{app:detection_results}

\subsubsection{Group-wise Detection at Different Localization Thresholds}
Table~\ref{tab:detection_group_thresholds} reports detection precision for each animal group at IoU thresholds of 0.50 and 0.75, showing performance under different localization requirements.

\begin{wraptable}{r}{0.61\textwidth}
    \centering
    \vspace{-0.2cm}
    \caption{AP$_{50}$ and AP$_{75}$ (\%) by animal group.}
    \label{tab:detection_group_thresholds}
    \footnotesize
    \setlength{\tabcolsep}{4pt}
    \setlength{\aboverulesep}{0pt}
    \setlength{\belowrulesep}{0pt}
    \renewcommand{\arraystretch}{1.1}
    \begin{adjustbox}{width=\linewidth}
    \begin{tabular}{llcccccc}
        \toprule
        \rowcolor{gray!12}
        Protocol & Metric
        & Mammals & Fish & Birds & Amph. & Rept. & Others \\
        \midrule
        \multirow{2}{*}{P1}
        & AP$_{50}$ & 85.28 & 82.57 & 87.71 & 86.01 & 29.76 & 62.03 \\
        & AP$_{75}$ & 77.15 & 77.04 & 79.60 & 77.31 & 26.31 & 47.70 \\
        \midrule
        \multirow{2}{*}{P2}
        & AP$_{50}$ & 69.15 & 80.84 & 89.73 & -- & -- & 33.82 \\
        & AP$_{75}$ & 57.04 & 72.38 & 77.86 & -- & -- & 23.33 \\
        \bottomrule
    \end{tabular}
\end{adjustbox}
\end{wraptable}

\subsubsection{Average Recall at Different Prediction Limits}

We complement the precision analysis in Table~\ref{tab:detection_overall} with average recall at different prediction limits. Evaluation is category-agnostic, mapping all valid animals to the \texttt{animal} foreground class. Table~\ref{tab:detection_recall} reports average recall with at most 1, 10, and 100 predicted boxes per frame, averaged over IoU thresholds from 0.50 to 0.95. All results are percentages.

\begin{wraptable}{r}{0.45\textwidth}
    \centering
    \footnotesize
    \vspace{-0.2cm}
    \caption{Average recall at different per-frame prediction limits under both protocols.}
    \label{tab:detection_recall}
    \centering
    \setlength{\aboverulesep}{0pt}
    \setlength{\belowrulesep}{0pt}
    \renewcommand{\arraystretch}{1.1}
    \begin{adjustbox}{width=\linewidth}
    \begin{tabular}{l ccc}
        \toprule
        \rowcolor{gray!12}
        Protocol & $\text{AR}@1$ & $\text{AR}@10$ & $\text{AR}@100$ \\
        \midrule
        Protocol 1 & 42.40 & 79.67 & 80.45 \\
        Protocol 2 & 35.63 & 69.38 & 70.09 \\
        \bottomrule
    \end{tabular}
    \end{adjustbox}
\end{wraptable}

Increasing the prediction limit from 1 to 10 improves average recall by 37.27 and 33.75 percentage points under the two protocols, reflecting the need for multiple candidates in multi-instance scenes. Raising the limit to 100 yields only 0.78 and 0.71 additional points, indicating limited aggregate recall gains from retaining more candidates under the current detector outputs and test distributions.

\subsection{Tracking Results by Animal Group}
\label{app:group_tracking}

Table~\ref{tab:group_hota} reports tracking performance by animal group. In Protocol~2, all methods achieve higher HOTA on birds than on the ``other animals'' group, revealing substantial variation beyond overall scores. Together with the detection analysis in the main text, these results characterize performance differences across diverse animal categories.

\begin{wraptable}{r}{0.65\textwidth}
    \centering
    \vspace{-0.2cm}
    \caption{Average HOTA by coarse animal group under Protocol~1 and Protocol~2. Groups follow the definition in Table~\ref{tab:detection_overall}.}
    \label{tab:group_hota}
    \centering
    \footnotesize
    \setlength{\aboverulesep}{0pt}
    \setlength{\belowrulesep}{0pt}
    \renewcommand{\arraystretch}{1.1}
    \begin{adjustbox}{max width=\linewidth}
    \begin{tabular}{lcccccc}
        \toprule
        \rowcolor{gray!12}
        Method & Mammals & Fish & Birds & Amphibians & Reptiles & Others \\
        \midrule
        \multicolumn{7}{c}{\textit{Protocol 1}} \\
        \midrule
        DiffMOT    & 67.02 & 63.72 & 71.97 & 83.98 & 49.42 & 50.02 \\
        TrackTrack & 66.05 & 62.58 & 69.96 & 83.17 & 48.48 & 50.37 \\
        MOTIP      & 63.58 & 57.00 & 68.28 & 78.21 & 51.82 & 49.91 \\
        TCEI       & 62.95 & 56.35 & 68.90 & 80.68 & 49.22 & 51.97 \\
        ByteTrack  & 57.65 & 51.05 & 60.50 & 80.44 & 42.42 & 47.45 \\
        QDTrack    & 54.73 & 50.13 & 58.35 & 72.09 & 35.92 & 35.22 \\
        TransTrack & 42.14 & 35.70 & 46.70 & 64.29 & 23.83 & 17.69 \\
        FairMOT    & 41.13 & 34.19 & 42.13 & 66.86 & 23.74 & 29.51 \\
        \midrule
        \multicolumn{7}{c}{\textit{Protocol 2}} \\
        \midrule
        DiffMOT    & 54.41 & 54.92 & 70.59 & -- & -- & 25.36 \\
        TrackTrack & 54.18 & 53.42 & 68.22 & -- & -- & 26.73 \\
        TCEI       & 51.79 & 49.80 & 68.09 & -- & -- & 34.85 \\
        MOTIP      & 49.72 & 49.41 & 70.31 & -- & -- & 33.94 \\
        ByteTrack  & 45.92 & 43.37 & 58.32 & -- & -- & 24.19 \\
        QDTrack    & 41.09 & 40.54 & 56.67 & -- & -- & 19.22 \\
        TransTrack & 33.24 & 32.61 & 49.97 & -- & -- & 13.95 \\
        FairMOT    & 29.38 & 26.99 & 43.46 & -- & -- & 18.11 \\
        \bottomrule
    \end{tabular}
    \end{adjustbox}
    \vspace{-0.6cm}
\end{wraptable}

\subsection{Per-Video HOTA and Stratified Statistics}

\label{app:video_hota_stats}

This subsection provides full statistics for the video-level long-tail analysis in Section~\ref{sec:evaluation_results}. Protocol~1 and Protocol~2 each contain 315 test videos but use different videos, precluding paired video-level comparisons.

\subsubsection{Complete Video-Level HOTA Statistics}

Table~\ref{tab:video_hota_stats_full} reports HOTA distributions for eight methods across the 315 test videos in each protocol. Mean is the arithmetic mean of per-video HOTA; Median, IQR, P10, and P90 denote the median, interquartile range, and 10th and 90th percentiles. The last two columns give the percentages of videos with HOTA below 20 and 40. These statistics weight videos equally, unlike the TrackEval aggregate results in the main comparison table.

\begin{figure}[H]
    \centering
    \appendixfigure{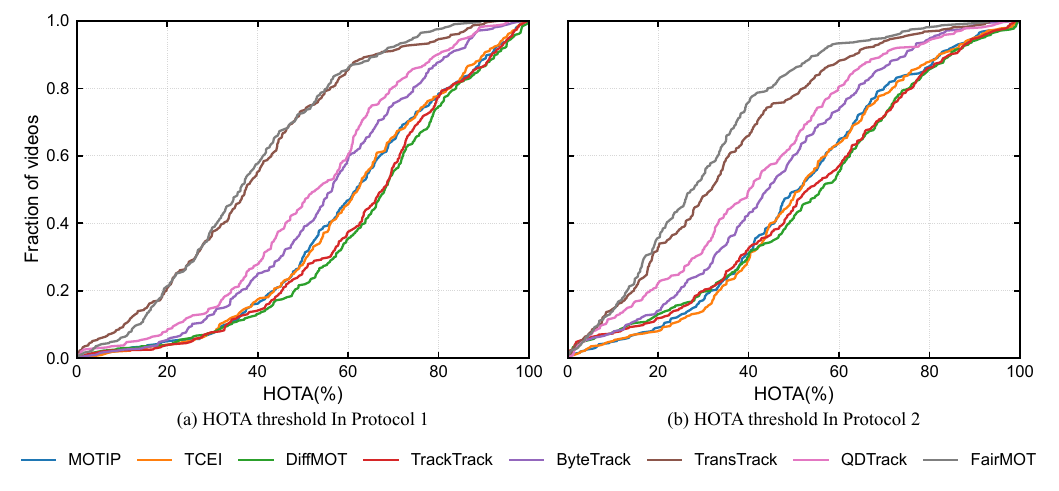}{0.42\textheight}
    \caption{Empirical CDF of per-video HOTA for all methods under Protocol~1 (left) and Protocol~2 (right). The curves show the fraction of videos below a given HOTA threshold.}
    \label{fig:video_hota_ecdf}
\end{figure}

In Protocol~2, DiffMOT has the highest mean per-video HOTA of 53.14. TCEI averages 51.95 but has a P10 of 23.18, exceeding DiffMOT's 14.31. Thus, average performance and performance on low-scoring videos provide complementary information. Figure~\ref{fig:video_hota_ecdf} shows the full empirical cumulative distributions of per-video HOTA.

\subsubsection{Grouping Videos by Performance}

Within each protocol, we compute each video's arithmetic mean HOTA across the eight baselines and sort videos in descending order. The top, middle, and bottom 105 videos form the Easy, Medium, and Hard groups, respectively. All methods share the same groups within a protocol, and Table~\ref{tab:difficulty_hota} reports their arithmetic mean HOTA in each group. These groups are defined by the current baseline set's test performance and are constructed separately because the protocols use different test videos.

\begin{table}[t]
    \centering
    \caption{Full video-level HOTA statistics on the 315 test videos of each protocol.}
    \label{tab:video_hota_stats_full}
    \centering
    \footnotesize
    \setlength{\aboverulesep}{0pt}
    \setlength{\belowrulesep}{0pt}
    \renewcommand{\arraystretch}{1.1}
    \begin{adjustbox}{max width=\linewidth}
    \begin{tabular}{lccccccc}
        \toprule
        \rowcolor{gray!12}
        Tracker & Mean & Median & IQR & P10 & P90 & \%HOTA$<$20 & \%HOTA$<$40 \\
        \midrule
        \multicolumn{8}{c}{\textit{Protocol 1}} \\
        \midrule
        TransTrack & 38.45 & 37.40 & 29.95 & 10.77 & 65.97 & 20.63 & 55.24 \\
        QDTrack    & 51.84 & 53.13 & 28.32 & 22.42 & 79.73 & 7.94  & 27.62 \\
        FairMOT    & 37.90 & 36.19 & 29.63 & 14.19 & 66.55 & 21.27 & 57.46 \\
        ByteTrack  & 55.22 & 56.46 & 29.34 & 26.28 & 82.25 & 5.08  & 25.08 \\
        DiffMOT    & 65.19 & 68.11 & 27.39 & 34.33 & 92.74 & 3.81  & 13.02 \\
        MOTIP      & 61.26 & 61.87 & 30.07 & 33.87 & 91.31 & 4.76  & 15.87 \\
        TrackTrack & 64.05 & 67.84 & 29.36 & 33.52 & 91.74 & 3.81  & 13.97 \\
        TCEI       & 61.27 & 62.15 & 28.84 & 31.94 & 89.98 & 3.81  & 17.14 \\
        \midrule
        \multicolumn{8}{c}{\textit{Protocol 2}} \\
        \midrule
        TransTrack & 33.39 & 31.91 & 27.54 & 7.48  & 63.51 & 32.70 & 65.71 \\
        QDTrack    & 40.62 & 40.15 & 31.56 & 8.53  & 69.66 & 22.22 & 49.52 \\
        FairMOT    & 29.91 & 27.86 & 23.74 & 6.86  & 55.35 & 35.56 & 75.56 \\
        ByteTrack  & 44.43 & 44.27 & 30.86 & 13.88 & 74.76 & 13.97 & 42.86 \\
        DiffMOT    & 53.14 & 55.95 & 35.39 & 14.31 & 84.77 & 13.02 & 30.16 \\
        MOTIP      & 51.27 & 51.10 & 29.60 & 21.33 & 82.34 & 8.89  & 30.79 \\
        TrackTrack & 52.47 & 53.12 & 36.96 & 17.22 & 84.50 & 11.75 & 32.38 \\
        TCEI       & 51.95 & 51.63 & 29.59 & 23.18 & 82.13 & 7.94  & 28.89 \\
        \bottomrule
    \end{tabular}
    \end{adjustbox}
\end{table}

\begin{table}[htbp]
    \centering
    \caption{Mean per-video HOTA on the Easy, Medium, and Hard groups.}
    \label{tab:difficulty_hota}
    \footnotesize
    \setlength{\aboverulesep}{0pt}
    \setlength{\belowrulesep}{0pt}
    \renewcommand{\arraystretch}{1.1}
    \begin{adjustbox}{width=\linewidth}
        \begin{tabular}{l cccccccc}
            \toprule
            \rowcolor{gray!12}
            Difficulty & DiffMOT & TrackTrack & MOTIP & TCEI & ByteTrack & QDTrack & FairMOT & TransTrack \\
            \midrule
            \multicolumn{9}{c}{\textit{Protocol 1}} \\
            \midrule
            Easy    & \textbf{86.03} & \underline{84.78} & 83.89 & 83.06 & 75.43 & 71.27 & 57.51 & 56.46 \\
            Medium  & \textbf{68.14} & \underline{67.21} & 61.37 & 62.30 & 56.81 & 54.16 & 37.36 & 37.93 \\
            Hard    & \textbf{41.39} & \underline{40.15} & 38.51 & 38.47 & 33.42 & 30.09 & 18.83 & 20.95 \\
            \midrule
            \multicolumn{9}{c}{\textit{Protocol 2}} \\
            \midrule
            Easy    & \textbf{79.11} & \underline{77.90} & 74.89 & 74.86 & 66.80 & 62.81 & 48.76 & 54.03 \\
            Medium  & \textbf{55.59} & \underline{55.43} & 50.64 & 51.91 & 45.63 & 41.54 & 28.53 & 31.19 \\
            Hard    & 24.73 & 24.09 & \underline{28.27} & \textbf{29.08} & 20.85 & 17.51 & 12.44 & 14.96 \\
            \bottomrule
        \end{tabular}
    \end{adjustbox}
\end{table}

Table~\ref{tab:difficulty_hota} shows that
DiffMOT achieves the highest mean HOTA in all three Protocol~1 groups. In Protocol~2, it leads the Easy and Medium groups, whereas TCEI and MOTIP achieve 29.08 and 28.27 on the Hard group, exceeding DiffMOT's 24.73. Relative strengths therefore vary across video groups, providing information beyond aggregate metrics.

\section{CDA Implementation and Additional Experiments}
\label{app:cda_details}

\begin{wraptable}{r}{0.5\textwidth}
    \centering
    \vspace{-0.3cm}
    \caption{CenterSim denominator ablation on the Protocol~2 test set ($\alpha=0.5$).}
    \label{tab:cda_denominator}
    \footnotesize
    \setlength{\tabcolsep}{4pt}
    \setlength{\aboverulesep}{0pt}
    \setlength{\belowrulesep}{0pt}
    \renewcommand{\arraystretch}{1.1}
    \begin{adjustbox}{width=\linewidth}
    \begin{tabular}{ccccc}
    \toprule
    \rowcolor{gray!12}
    Denominator & HOTA & AssA & IDF1 & IDSW \\
    \midrule
    mean (default)
    & 53.80 & 58.25 & 56.81 & 1972 \\
    max
    & 53.58 & 57.77 & 56.60 & 1992 \\
    \rowcolor{VastMAT}
    min
    & \textbf{53.82} & \textbf{58.32} & 56.62 & 1953 \\
    enclosing
    & 53.18 & 56.89 & 56.05 & 2018 \\
    \bottomrule
    \end{tabular}
    \end{adjustbox}
    \vspace{-1cm}
\end{wraptable}

This section details CDA's association cost and candidate gating, and examines its design through fusion-weight, density-parameter, and post-processing comparisons. Offline candidate-pair diagnostics characterize the complementarity of IoU and center similarity, while runtime measurements assess integration overhead.

\subsection{Association Cost and Candidate Gating}

\label{app:cda_gate}

CDA operates at inference time and requires no additional training. In TrackTrack, the association cost is
\begin{equation}
C(a,b)=0.50\bigl(1-S(a,b)\bigr)
+0.50D_{\mathrm{app}}(a,b)
+0.10D_{\mathrm{conf}}(a,b)
+0.05D_{\mathrm{angle}}(a,b),
\end{equation}
where $D_{\mathrm{app}}$, $D_{\mathrm{conf}}$,
and $D_{\mathrm{angle}}$
denote appearance cosine distance, confidence difference, and motion-direction difference, respectively; $S$ is the fused similarity in Equation~\eqref{eq:similarity}.

Original TrackTrack retains its original gating. The normalized-DIoU control and CDA variants share DIoU gating, rejecting a candidate pair when $\widetilde{\mathrm{DIoU}}(a,b)\leq0.10$. Among these controls, CDA changes association costs while preserving the gating function and threshold.

\begin{equation}
\widetilde{\mathrm{DIoU}}(a,b)=\mathrm{clip}\left(
\frac{
\mathrm{IoU}(a,b)-\min((d_{\mathrm{center}}/c)^2,1)+1}{2},\,0,\,1\right),
\label{eq:diou}
\end{equation}

\begin{wrapfigure}{r}{0.45\textwidth}
    \centering
    \vspace{-0.3cm}
    \appendixfigure{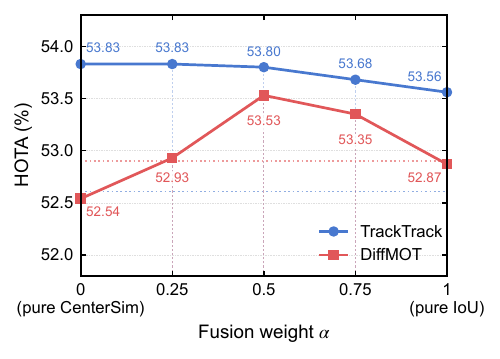}{0.49\textheight}
    \caption{Fusion-weight sensitivity of TrackTrack and DiffMOT.}
    \label{fig:alpha_curves}
    \vspace{-0.8cm}
\end{wrapfigure}

Here, $c$ is the diagonal length of the smallest enclosing rectangle. Association parameters are max\_time\_lost=20
det\_thr=init\_thr=match\_thr=0.60, tai\_thr=0.55, and penalty\_p/q=0.20/0.40, with original AFLink post-processing enabled by default. Apart from the explicitly reported geometric similarity and gating configurations, cached detections, FastReID features, TPA, TAI, and all other association parameters remain identical.

\subsection{Normalization Denominator Comparison}
\label{app:cda_normalization}

Table~\ref{tab:cda_denominator} fixes $\alpha=0.5$ and varies only the center-distance denominator, retaining the other settings from Section~\ref{sec:exp_cda}. It reports association metrics and identity-switch counts for each configuration.

\subsection{Fusion-Weight Sensitivity}
\label{app:fusion_weights}

\begin{wraptable}{r}{0.4\textwidth}
    \centering
    \vspace{-0.3cm}
    \footnotesize
    \setlength{\tabcolsep}{4pt}
    \setlength{\aboverulesep}{0pt}
    \setlength{\belowrulesep}{0pt}
    \renewcommand{\arraystretch}{1.1}
    \caption{TrackTrack fusion-weight sensitivity on Protocol~2.}
    \label{tab:cda_alpha}
    \begin{adjustbox}{max width=\linewidth}
    \begin{tabular}{cccc}
      \toprule
      \rowcolor{gray!12}
      $\alpha$ & HOTA & $\Delta$ vs TrackTrack & IDSW \\
      \midrule
      \rowcolor{VastMAT}
      0     & \textbf{53.83} & \textbf{+1.22} & 2313 \\
      0.25  & 53.83 & +1.22 & 2054 \\
      0.5   & 53.80 & +1.19 & 1972 \\
      0.75  & 53.68 & +1.07 & \textbf{1956} \\
      1     & 53.56 & +0.95 & 2082 \\
      \bottomrule
    \end{tabular}
    \end{adjustbox}
\end{wraptable}

Figure~\ref{fig:alpha_curves} compares TrackTrack and DiffMOT across fusion weights, with numerical results in Tables~\ref{tab:cda_alpha}
and~\ref{tab:diffmot_alpha}. TrackTrack uses identical DIoU gating and AFLink settings across weights. As $\alpha$ increases from 0 to 1, HOTA ranges from 53.56 to 53.83;
$\alpha=0$ and $\alpha=0.25$ both achieve 53.83, but the latter reduces IDSW from 2313 to 2054. Increasing the IoU weight to $\alpha=0.75$ further reduces IDSW to 1956, with 53.68 HOTA, illustrating the differing effects of fusion weights on overall association and identity switches.

\begin{wraptable}{r}{0.5\textwidth}
    \centering
    \vspace{-0.3cm}
    \footnotesize
    \setlength{\aboverulesep}{0pt}
    \setlength{\belowrulesep}{0pt}
    \renewcommand{\arraystretch}{1.1}
    \caption{DiffMOT fusion-weight sensitivity on Protocol~2.}
    \label{tab:diffmot_alpha}
    \begin{adjustbox}{max width=\linewidth}
    \begin{tabular}{ccccc}
      \toprule
      \rowcolor{gray!12}
      $\alpha$ & HOTA & AssA & IDSW & $\Delta$HOTA \\
      \midrule
      baseline (CDA off)  & 52.90 & 57.61 & 2545 & -- \\
      $0$ & 52.54 & 56.44 & 3188 & $-0.36$ \\
      $0.25$ & 52.93 & 57.30 & 2711 & $+0.03$ \\
      \rowcolor{VastMAT}
      $0.5$  & \textbf{53.53} & \textbf{58.63} & 2467 & \textbf{$+0.63$} \\
      $0.75$ & 53.35 & 58.35 & \textbf{2439} & $+0.45$ \\
      $1$    & 52.87 & 57.54 & 2528 & $-0.03$ \\
      \bottomrule
    \end{tabular}
    \end{adjustbox}
\end{wraptable}

DiffMOT achieves its highest HOTA of 53.53 at $\alpha=0.5$, improving on the baseline by 0.63 percentage points
and outperforming both pure center similarity and pure IoU. The trackers' responses indicate that overlap and center proximity are complementary, with their effective balance depending on the base tracker's motion prediction and association configuration.

\subsection{Additional CDA Analyses}
\label{app:cda_extra}

\begin{wraptable}{r}{0.5\textwidth}
    \centering
    \vspace{-0.3cm}
    \footnotesize
    \setlength{\tabcolsep}{4pt}
    \setlength{\aboverulesep}{0pt}
    \setlength{\belowrulesep}{0pt}
    \renewcommand{\arraystretch}{1.1}
    \caption{AFLink sensitivity of two CDA configurations (Protocol~2).}
    \label{tab:cda_aflink}
    \begin{adjustbox}{max width=\linewidth}
    \begin{tabular}{cccccc}
      \toprule
      \rowcolor{gray!12}
      Config. & AFLink & HOTA & AssA & IDF1 & IDSW \\
      \midrule
      \rowcolor{VastMAT}
      Fixed $\alpha=0.25$ & On  & \textbf{53.83} & 58.19 & 56.89 & 2054 \\
      Fixed $\alpha=0.25$ & Off & 53.74 & 57.97 & 56.71 & 2068 \\
      \rowcolor{VastMAT}
      Adaptive            & On  & \textbf{53.92} & 58.38 & 57.06 & 2166 \\
      Adaptive            & Off & 53.79 & 58.11 & 56.83 & 2183 \\
      \bottomrule
    \end{tabular}
    \end{adjustbox}
\end{wraptable}

\paragraph{Sensitivity to AFLink Post-processing.}
Table~\ref{tab:cda_aflink} compares fixed-weight and adaptive CDA with and without AFLink.
Without AFLink, their HOTA scores are 53.74 and 53.79, respectively; enabling it increases them to 53.83 and 53.92. AFLink provides modest gains for both configurations, and adaptive CDA achieves higher HOTA in both post-processing settings.

\begin{wraptable}{r}{0.32\textwidth}
    \centering
    \vspace{-0.1cm}
    \footnotesize
    \setlength{\tabcolsep}{3pt}
    \setlength{\aboverulesep}{0pt}
    \setlength{\belowrulesep}{0pt}
    \renewcommand{\arraystretch}{1.1}
    \caption{Density-adaptive parameter grid on Protocol~2.}
    \label{tab:cda_density_grid}
    \begin{adjustbox}{max width=\linewidth}
    \begin{tabular}{lcccc}
      \toprule
      \rowcolor{gray!12}
      $k$ & $N_{\mathrm{ref}}$ & HOTA & AssA & IDSW \\
      \midrule
      0.02 & 5  & 53.47 & 57.53 & 2035 \\
      0.02 & 10 & 53.54 & 57.58 & 2041 \\
      0.02 & 20 & 53.82 & 58.17 & 2136 \\
      0.05 & 5  & 53.36 & 57.32 & 2034 \\
      \rowcolor{VastMAT}
      \textbf{0.05} & \textbf{10} & \textbf{53.92} & \textbf{58.38} & 2166 \\
      0.05 & 20 & 53.89 & 58.29 & 2171 \\
      0.10 & 5  & 53.49 & 57.53 & 2066 \\
      0.10 & 10 & 53.82 & 58.17 & 2171 \\
      0.10 & 20 & 53.78 & 58.05 & 2175 \\
      \bottomrule
    \end{tabular}
    \end{adjustbox}
    \vspace{-0.3cm}
\end{wraptable}

\paragraph{Density-Adaptive Parameter Grid.}
We vary the slope $k$ and reference detection count $N_{\mathrm{ref}}$ in Equation~\eqref{eq:alpha} to assess parameter sensitivity. Across the nine configurations in Table~\ref{tab:cda_density_grid}, HOTA ranges from 53.36 to 53.92, consistently above original TrackTrack's 52.61. The defaults $k=0.05$ and $N_{\mathrm{ref}}=10$, selected on the training-side development set and fixed for testing, achieve 53.92 HOTA and 58.38 AssA. These results characterize association performance across different adaptation rates and reference counts.

\paragraph{Additional Seen-Category Results.}
Table~\ref{tab:cda_protocol1} reports full Protocol~1 results for original TrackTrack and fixed-weight and adaptive CDA. The fixed configuration uses $\alpha=0.25$, while the adaptive configuration follows the fusion rule defined in the main text.

\begin{wraptable}{r}{0.6\textwidth}
  \centering
  \vspace{-0.2cm}
  \footnotesize
  \caption{Candidate-pair discrimination by normalized displacement (center distance divided by the mean box diagonal). Fractions indicate pairs correctly distinguished by only CenterSim or only IoU.}
  \label{tab:cda_speed}
  \setlength{\aboverulesep}{0pt}
  \setlength{\belowrulesep}{0pt}
  \renewcommand{\arraystretch}{1.1}
  \begin{adjustbox}{width=\linewidth}
    \begin{tabular}{cccc}
      \toprule
      \rowcolor{gray!12}
      Norm.\ displ. & only\_center & only\_iou & Better cue \\
      \midrule
      $\leq 0.25$    & 0.24\%           & 1.12\%         & IoU \\
      $0.25$--$0.5$  & 1.83\%           & 4.96\%         & IoU \\
      $0.5$--$1.0$   & \textbf{11.01\%} & 4.83\%         & \textbf{CenterSim ($2.3\times$)} \\
      $> 1.0$        & \textbf{5.94\%}  & 2.31\%         & \textbf{CenterSim ($2.6\times$)} \\
      \bottomrule
    \end{tabular}
  \end{adjustbox}
\end{wraptable}

\paragraph{Offline Candidate-Pair Diagnostics.}
To examine the discriminative complementarity of CenterSim and IoU, we analyze 190,890 association events offline over the full test set. For each event with a correct candidate, we define $m_I = \mathrm{IoU}(g_{\mathrm{prev}}, c^*) - \mathrm{IoU}(g_{\mathrm{prev}}, w^*)$ and $m_C = \mathrm{CenterSim}(g_{\mathrm{prev}}, c^*) - \mathrm{CenterSim}(g_{\mathrm{prev}}, w^*)$, where $g_{\mathrm{prev}}$ is the track's previous-frame GT box, $c^*$ is the correct candidate with the highest IoU of at least 0.5 with the current GT box, and $w^*$ is the strongest incorrect competitor, defined as the candidate other than $c^*$ with the highest IoU with $g_{\mathrm{prev}}$.

\begin{wraptable}{r}{0.58\textwidth}
    \centering
    \footnotesize
    \vspace{-0.2cm}
    \caption{CDA results on Protocol~1.}
    \label{tab:cda_protocol1}
    \setlength{\aboverulesep}{0pt}
    \setlength{\belowrulesep}{0pt}
    \renewcommand{\arraystretch}{1.1}
    \begin{adjustbox}{max width=\linewidth}
    \begin{tabular}{ccccc}
        \toprule
        \rowcolor{gray!12}
        Configuration & HOTA & AssA & IDF1 & IDSW \\
        \midrule
        TrackTrack (HMIoU)         & 65.10 & 65.26 & 67.87 & 1962 \\
        CDA ($\alpha=0.25$)              & 66.66 & 67.46 & 70.41 & 1662 \\
        \rowcolor{VastMAT}
        CDA (adaptive) & \textbf{66.68} & \textbf{67.49} & \textbf{70.49} & \textbf{1652} \\
        \bottomrule
    \end{tabular}
    \end{adjustbox}
\end{wraptable}

Table~\ref{tab:cda_speed} reports candidate-pair discrimination by normalized displacement. For displacements from $0.5$ to $1.0$, only CenterSim correctly distinguishes the selected pair in 11.01\% of events, compared with 4.83\% for only IoU. This diagnostic reveals different responses to low-overlap events. It measures discrimination between specified candidate pairs, while online tracking evaluates association over complete candidate sets.

\begin{wraptable}{r}{0.6\textwidth}
  \centering
  \vspace{-0.2cm}
  \footnotesize
  \setlength{\tabcolsep}{6pt}
  \caption{CDA inference overhead. The microbenchmark measures motion-similarity
    computation for one association call ($\mu$s, single-threaded).}
  \label{tab:inference_cost}
  \setlength{\aboverulesep}{0pt}
  \setlength{\belowrulesep}{0pt}
  \renewcommand{\arraystretch}{1.1}
  \begin{adjustbox}{width=\linewidth}
    \begin{tabular}{cccc}
      \toprule
      \rowcolor{gray!12}
      Association matrix & Pure IoU ($\mu$s) & CDA ($\mu$s) & Increase \\
      \midrule
      $20\times 40$   & 561.6   & 595.8   & $+34.2$ ($+6\%$) \\
      $50\times 100$  & 3419.9  & 3534.3  & $+114.4$ ($+3\%$) \\
      $100\times 200$ & 13256.7 & 13614.7 & $+358.0$ ($+3\%$) \\
      \bottomrule
    \end{tabular}
  \end{adjustbox}
\end{wraptable}

\paragraph{Inference Cost.}
Table~\ref{tab:inference_cost} reports the computational overhead of CDA's geometric similarity. With detections and appearance features cached, processing 93,111 frames takes 129.32\,s, compared with 128.97\,s for the baseline, an increase of 0.27\%. The microbenchmark isolates motion-similarity computation for a single association call.

\section{Dataset Documentation and Maintenance}
\label{app:maintenance_ethics}

\paragraph{Data sources and release.}
VastMAT is built from publicly available videos. Release materials will record sequence provenance, clip time intervals, and applicable licenses or permissions, providing media files or annotations with source indices as permitted. Code, annotations, and third-party videos will have separately stated licenses and attribution requirements.

\paragraph{Version maintenance.}
Each release will specify data and annotation versions, video lists, and evaluation configurations. Version histories will document annotation corrections, sequence removals, and split changes. If source videos become unavailable, affected sequences will be recorded; evaluations on the remaining subset will state their actual scope.

\paragraph{Issue reporting.}
The project will provide a channel for annotation errors, access issues, and copyright or privacy concerns. Verified issues will prompt corrections, access restrictions, or removal, with effects on the data and evaluation scope documented.

\paragraph{Intended use.}
VastMAT supports research on animal detection, within-video identity association, and Cross-category generalization. Its public-video sources introduce biases in category and scene distributions. Users should interpret aggregate metrics alongside category coverage, animal-group results, and video-level performance, and further validate models in their intended application environments.

\end{document}